\documentclass{article}
\usepackage{preprint,times}
\usepackage[T1]{fontenc}
\usepackage[utf8]{inputenc}
\usepackage{amsmath,amssymb,color}
\usepackage[hidelinks]{hyperref}
\usepackage{url,graphicx}
\usepackage{tikz}
\usepackage{xcolor} 
\colorlet{red}{black}
\definecolor{findingbackground}{rgb}{0.95,0.98,0.99}
\definecolor{findingborder}{rgb}{0.08,0.45,0.51}
\newsavebox{\findingcontent}
\newenvironment{findingbox}{%
  \par\medskip\begin{lrbox}{\findingcontent}%
  \begin{minipage}{\dimexpr\linewidth-16pt\relax}%
}{%
  \end{minipage}\end{lrbox}%
  \noindent\begin{tikzpicture}
    \node[fill=findingbackground,
      inner xsep=8pt,inner ysep=5pt,outer sep=0pt] (finding)
      {\usebox{\findingcontent}};
    \fill[findingborder] (finding.north west)
      rectangle ([xshift=2pt]finding.south west);
  \end{tikzpicture}\par\medskip
}
\title{Do Neural PDE Solvers Learn the Right Dynamics?}
\author{%
\parbox{0.97\textwidth}{\centering
\textbf{Haonan Li}$^{1}$, \textbf{Yue Song}$^{2}$\thanks{Denotes corresponding author.},
\textbf{Bin Yang}$^{1}$, \textbf{Kaihong Luo}$^{1,3,*}$\\[2pt]
{\normalfont\small
$^{1}$Center for Combustion Energy, Tsinghua University\quad
$^{2}$College of AI, Tsinghua University\\
$^{3}$Department of Mechanical Engineering, University College London
}\vspace{-10pt}
}}

\begin{document}
\addtocontents{toc}{\protect\setcounter{tocdepth}{-1}}
\maketitle

\begin{abstract}
Neural PDE solvers can achieve low prediction errors, but do they reproduce the dynamics of the systems they model? Prediction scores alone offer an incomplete answer: they measure agreement with reference solutions but provide limited insight into how errors accumulate, nearby states diverge, or extreme events arise. We propose an evaluation framework that directly examines these behaviors in deterministic and stochastic neural solvers. By evolving ensembles of nearby initial states and comparing them with direct numerical simulation, we assess three complementary aspects of learned dynamics: \textbf{error formation, ensemble geometry, and extreme events}. Experiments on two-dimensional Kolmogorov flow reveal limitations that conventional scores can obscure. Smaller trajectory errors can reflect weaker error amplification despite less accurate local updates. Models can match an ensemble's overall spread and effective dimension while failing to capture the spatial directions where nearby states diverge. Similarly, matching overall event frequencies can conceal failures to predict persistent extreme events. \emph{These findings show that improved prediction accuracy does not necessarily imply greater dynamical fidelity.} Our framework makes this distinction measurable, providing concrete criteria for evaluating whether advances in neural PDE solvers better capture the underlying dynamics.
\end{abstract}

\begin{figure}[!htbp]
\centering
\includegraphics[width=\linewidth]{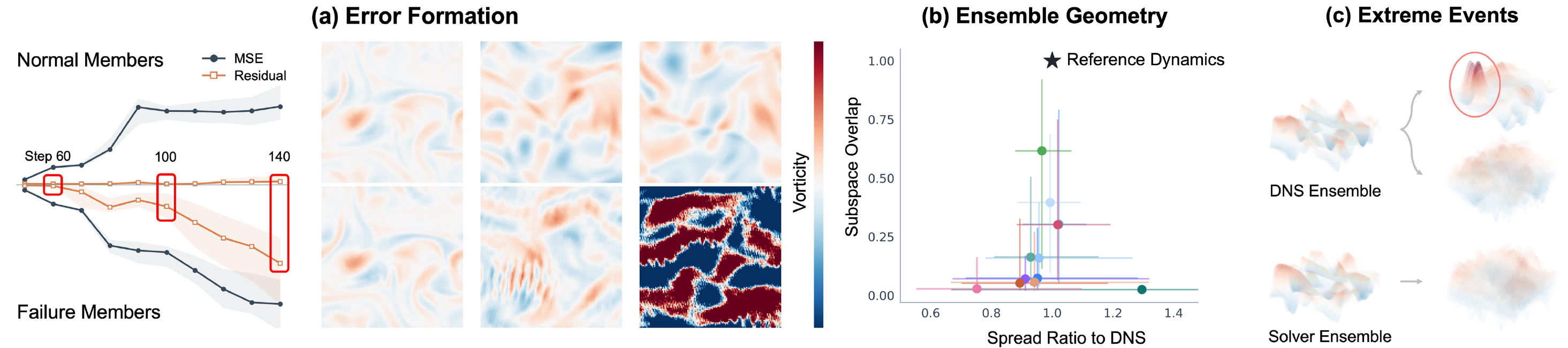}
\caption{\textbf{Probing learned dynamics beyond prediction error.} Ensembles of nearby initial states expose three distinct failures in neural solvers of two-dimensional Kolmogorov flow. \textbf{(a) Error formation:} larger one-step dynamical errors distinguish members that later fail even while their trajectory errors remain similar to those of non-failing members. The three columns of vorticity fields on the right correspond to the boxed forecast steps (60, 100, and 140) on the left and show their subsequent divergence. \textbf{(b) Ensemble geometry:} an ensemble spread close to DNS can coexist with low overlap between the dominant solver and reference subspaces. \textbf{(c) Extreme events:} nearby initial states can lead to distinct event and non-event outcomes under the reference dynamics, while a neural solver can miss the event-producing branch.}
\label{fig:teaser}
\vspace{3pt}
\end{figure}

\section{Introduction}

Neural partial differential equation (PDE) solvers have shown considerable success in predicting the evolution of physical systems. Recent pretrained models, including MPP, DPOT, and Poseidon, extend this capability across equations \citep{mccabe2024multiple,hao2024dpot,herde2024poseidon}. Progress is primarily measured through pointwise field errors, with studies of chaotic systems and generative models also employing distributional distances and probabilistic scores \citep{schiff2024dyslim,molinaro2024generative,ruhling2023dyffusion}. These evaluations quantify agreement with reference solutions, but leave a fundamental unsolved question: 

\begin{findingbox}
\centering
\emph{Do neural PDE solvers learn the right dynamics?}
\end{findingbox}

Prediction accuracy alone provides an incomplete answer. During a rollout, each model update introduces errors that interact with discrepancies inherited from earlier steps. A smaller trajectory error may therefore reflect more accurate updates, weaker amplification of existing errors, or cancellation between the two. Beyond individual trajectories, the dynamics determine how nearby initial states separate and which spatial patterns dominate their evolution. These responses are not established by matching an ensemble's overall spread. Similarly, agreement in aggregate event frequency can conceal errors under particular initial conditions, while aggregate distributional scores may inadequately expose relative errors in rare-event probabilities \citep{gneiting2011comparing}. Understanding learned dynamics thus requires examining how predictions evolve and which properties they preserve.

In this work, we investigate three questions: \emph{How do model errors form and propagate? Do nearby initial states evolve along the same directions as in the reference system? Do extreme events occur with the right intensities and under the right initial conditions?} We study these questions through a common experimental protocol: densely sample perturbed initial states around the initial state of each case, evolve each ensemble using both the neural solver and direct numerical simulation (DNS), and compare their evolution. These perturbations serve as controlled probes of the learned dynamics; they do not presuppose uncertainty in the initial conditions. The protocol accommodates deterministic and stochastic solvers. For stochastic solvers, ensemble members share a random path, with fresh randomness at successive steps, so that differences across members reflect their responses to initial-state perturbations under the same realization of model randomness.

Our experiments on deterministic chaotic two-dimensional Kolmogorov flow identify three discrepancies between predictive performance and dynamical fidelity.

\noindent\textbf{Error formation: lower trajectory error need not imply more accurate local updates.}
We decompose error at successive updates into newly introduced model residuals, propagation of existing discrepancies, and their interaction. The analysis reveals cases in which persistently larger local residuals coexist with lower trajectory errors because existing discrepancies undergo weaker dynamical amplification. The residuals also alter the collective evolution of nearby states: early in prediction, they reduce differences between ensemble members relative to DNS advancement from the same model states, while increasing the error of the centered ensemble against the DNS reference. Thus, keeping trajectories closer together can accompany a less faithful response to initial-state perturbations.

\noindent\textbf{Ensemble geometry: matching ensemble spread and dimension does not ensure the correct spatial directions.} As the DNS ensemble evolves, its variance first concentrates into a few directions and later spreads across more directions, while the dominant directions themselves change. Models reproduce the broad evolution of total variance and effective dimension, yet most tested configurations show substantial misalignment with the dominant DNS subspace at intermediate prediction times. This discrepancy persists even when variance and effective dimension closely match DNS. Capturing how much an ensemble spreads and how many directions it occupies therefore does not establish that it evolves along the correct directions.

\noindent\textbf{Extreme events: correct aggregate event frequencies can conceal incorrect conditional probabilities.} Models whose overall extreme-event frequencies closely match DNS can substantially overestimate or underestimate event probabilities within individual initial-state ensembles. These errors cancel upon aggregation, allowing apparently correct overall frequencies to conceal failures to predict extreme events that persist under the reference dynamics. Even when predicted vortices remain recognizable and have similar peak intensities, their deformation and rotation can differ substantially from the reference dynamics.

Together, these findings show how favorable prediction scores can coexist with systematic errors in learned dynamics. Our investigation complements conventional evaluation by tracing error formation, testing the geometry of nearby trajectories, and resolving extreme-event statistics by initial condition. These analyses provide concrete criteria for assessing whether improvements in neural PDE solvers correspond to a more faithful representation of the underlying system.

\section{Experimental Setup and Dynamical Probes}
\label{sec:setup}

We probe learned dynamics by evolving ensembles of nearby initial states with both direct numerical simulation (DNS) and neural solvers. This allows us to examine individual prediction errors alongside the collective response to initial-state perturbations. All three investigations share the same initial states, data split, and prediction origin.

\noindent\textbf{Reference Dynamics.} We consider two-dimensional Kolmogorov flow, governed by the incompressible Navier--Stokes equations with sinusoidal forcing and periodic boundaries. Although deterministic, this system exhibits chaotic evolution and develops localized intense vortices \citep{vela2024large}.

\noindent\textbf{Dataset Construction and Initial-State Probes.} We sample temporally decorrelated base states from the stationary-flow archive of \citet{vela2024large}, with each state defining a case $c$. Spatially structured perturbations, constructed from rescaled, randomly signed differences between archive states, define nearby ensemble members $m$. We denote their vorticity fields by $x^q_{c,m}(t)\in\mathbb{R}^D$, where $q$ identifies the solver and $D$ is the spatial dimension. Cases are split into training, validation, and test sets. Training and validation use only unperturbed trajectories; ensemble diagnostics use perturbed test members, each with a matched DNS reference. Appendix~\ref{app:setup} details the numerical settings, dataset sizes, forecast initialization, and stochastic sampling protocol.

\noindent\textbf{Solvers and Forecast Evaluation.} We evaluate configurations of U-Net \citep{ronneberger2015u}, FNO \citep{li2020fourier}, AFNO \citep{guibas2021adaptive}, DPOT \citep{hao2024dpot}, DySLIM \citep{schiff2024dyslim}, PDE-Refiner \citep{lippe2023pde}, ACDM \citep{kohl2026benchmarking}, PFNO, and DYffusion \citep{ruhling2023dyffusion}. Configuration labels specify the input history length $H$, output block length $P$, and training unroll length $K$. We exclude configurations that perform poorly on conventional forecast metrics from dynamical evaluation; Appendix~\ref{app:models} provides the screening details.

\section{Does Lower \textcolor{red}{Rollout Error} Imply More Accurate Local Updates?}
\label{sec:errors}

\textcolor{red}{Rollour Prediciton Error} couples \textcolor{red}{local prediciton} with the \textcolor{red}{dynamical propagation} of existing discrepancies. We first separate these contributions to examine whether lower forecast error reflects more accurate local updates. We then ask whether training that improves forecast accuracy also preserves the response to nearby initial states.

\subsection{Lower \textcolor{red}{Rollout Prediction Error} Can \textcolor{red}{Obscure} Larger \textcolor{red}{Local Prediction Error}}

{\color{red}
Let $x_k$ denote the DNS ground truth state at step $k$ and $\hat{x}_k$ denote the state predicted by the neural solver. Let $F$ denote the one-step map for DNS and $\hat{F}$ for the neural solver, so that $x_{k+1} = F(x_k),   \hat{x}_{k+1} = \hat{F}(\hat{x}_k).$ We define the \textit{rollout prediction error} at step k as $e_k = \hat{x}_k - x_k$. 

At each step $k$, we apply the DNS map $F$ to the predicted state $\hat{x}_k$ to obtain $F(\hat{x}_k)$. Then we give the following decomposition of the $e_{k+1}$:}
\begin{equation}
    e_{k+1} = \hat{x}_{k+1} - x_{k+1}
    = \underbrace{\bigl(\hat{x}_{k+1} - F(\hat{x}_k)\bigr)}_{r_{k+1}}
    + \underbrace{\bigl(F(\hat{x}_k) - x_{k+1}\bigr)}_{h_{k+1}}.
\end{equation}

{\color{red}
$r_{k+1}$ denotes \textit{local prediction error}, which is the one-step map difference between the neural solver $\hat{F}$ and the DNS $F$ starting from the same $\hat{x}_k$. $h_{k+1}$ denotes \textit{propagated existing error}, which tells how the existing error between $\hat{x}_k$ and $x_k$ propagtes under the ground truth dynamics.}

{\color{red}
The change in squared rollout prediction error from step $k$ to $k+1$
is therefore:}
\begin{equation}
    \|e_{k+1}\|^2 - \|e_k\|^2
    =
    \underbrace{\bigl(\|h_{k+1}\|^2 - \|e_k\|^2\bigr)}_{\mathbf{g}_{k+1}}
    + \underbrace{\|r_{k+1}\|^2}_{\mathbf{j}_{k+1}}
    + \underbrace{2\langle h_{k+1}, r_{k+1}\rangle}_{\mathbf{c}_{k+1}}.
\end{equation}

The three terms quantify, \textcolor{red}{respectively,} the propagation of existing error \textcolor{red}{by the ground truth dynamics ($g_{k+1}$)}, the squared \textcolor{red}{local prediction error ($j_{k+1}$)}, and the directional coupling between \textcolor{red}{$h_{k+1}$ and $r_{k+1}$ ($c_{k+1}$).} {\color{red} Summing over the first \(k\) steps gives three time-aggregated terms,
\(G\), \(J\), and \(C\), which can be directly compared with the squared error:
\begin{equation}
    \|e_k\|^2
    =
    \underbrace{\sum_{i=1}^{k} g_i}_{G}
    + \underbrace{\sum_{i=1}^{k} j_i}_{J}
    + \underbrace{\sum_{i=1}^{k} c_i}_{C}
    = G + J + C.
\end{equation}}

\begin{figure}[!t]
\centering
\includegraphics[width=\linewidth]{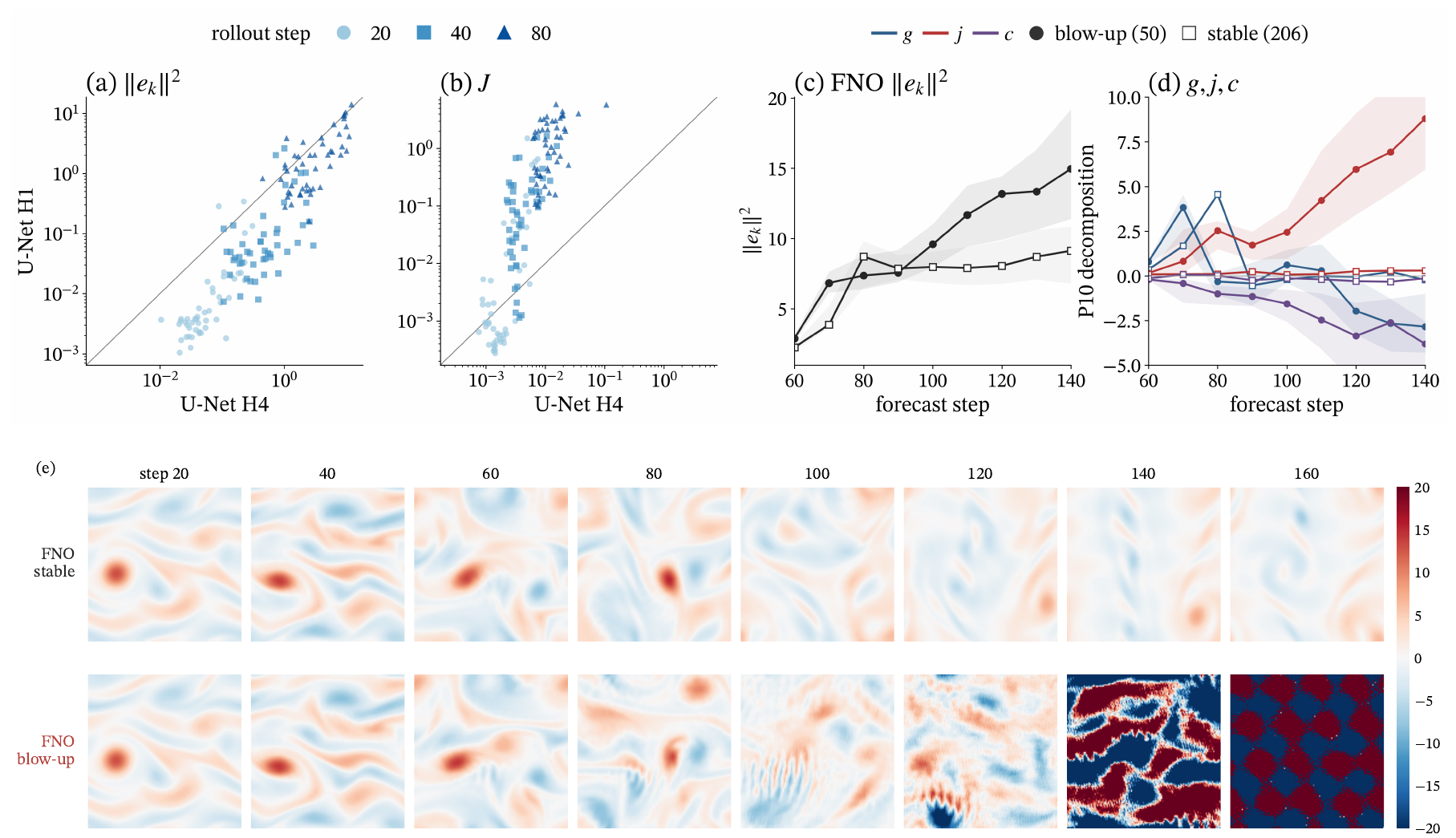}
\caption{\emph{\textcolor{red}{Local prediction error} can remain hidden in \textcolor{red}{rollout prediction error} and precede rollout failure.} \textbf{(a--b)} \textcolor{red}{$\|e_k\|^2$} and \textcolor{red}{$J$} for U-Net H1 vs. H4 \textcolor{red}{($H_i$ indicates that the training input uses the past $i$ frames).} Each point represents one initial condition and reports the average over all members under that initial condition. \textbf{(c--d)} \textcolor{red}{$\|e_k\|^2$} and $g$, $j$, and $c$ for 50 blow-up and 206 stable FNO trajectories in case 4232. Curves show medians; shading shows interquartile ranges. \textbf{(e)} Vorticity fields of a \textcolor{red}{stable} and a \textcolor{red}{blow-up} trajectory from the same case.}
\label{fig:budget}
\vspace{3pt}
\end{figure}

\textcolor{red}{Lower $\|e_k\|^2$ can obscure larger $J$, due to small or negative $G$ and $C$. A good solver should first and foremost have good local accuracy. However, the rollout prediction error, as a coupled observation, can mask poor local predictions when they occur.} As shown in Figure~\ref{fig:budget}a--b, under almost all initial conditions, U-Net H1 achieves lower \textcolor{red}{rollout prediction error $\|e_k\|^2$} than H4 despite larger cumulative \textcolor{red}{local prediction error $J$.}

\textcolor{red}{The obscuring effect can be harmful for long-horizon rollout stability. Despite having the smallest \(\|e_k\|^2\) in the first 100 steps, FNO exhibits numerical blow-up after step 140. Member failures occur in 3 of the 48 test cases, peaking at 50 of 256 members in case 4232 (Figure~\ref{fig:budget}e). For all 50 blown-up members,} $\|e_k\|^2$ during steps 60--100 is almost identical to that of the stable members. In contrast, their local prediction error \textcolor{red}{$j$} is consistently and significantly higher.

\begin{figure}[!t]
\renewcommand{\thefigure}{R1} 
\centering
\includegraphics[width=\linewidth]{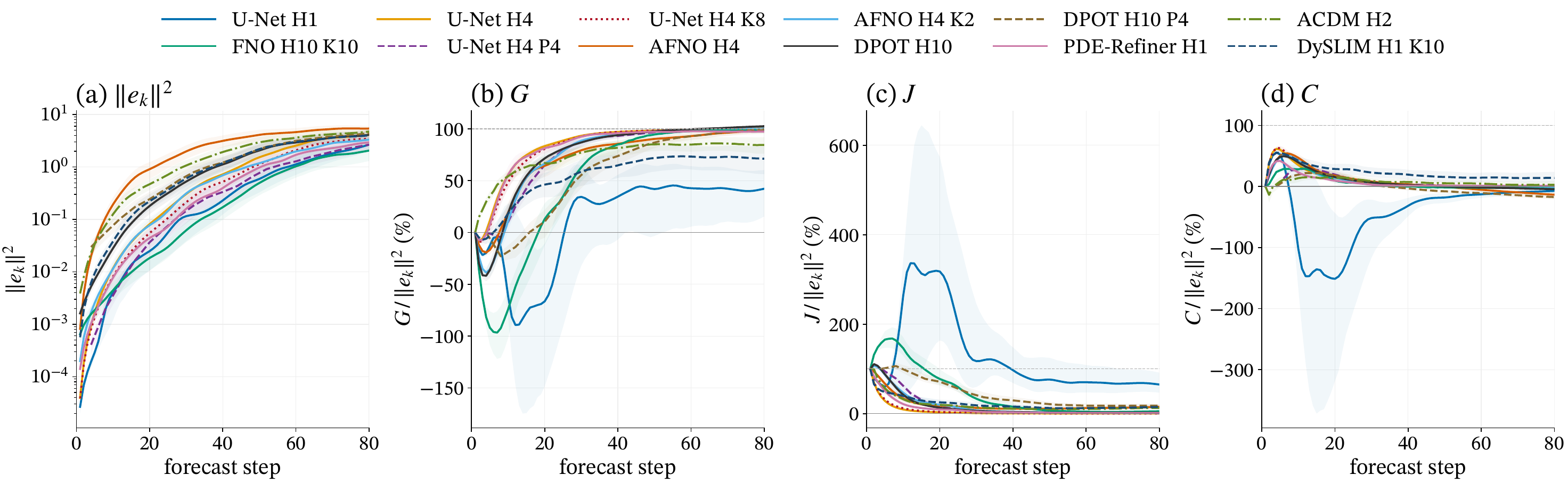}
\caption{\textcolor{red}{\textbf{(a)} $\|e_k\|^2$. \textbf{(b--d)} Percentage of $G$, $J$, $C$. Note that $\|e_k\|^2 = G + J + C.$ Lines are equal-case means over 48 cases (ratios of means in (b--d)); bands are pointwise $90\%$ case-bootstrap confidence intervals from 2,000 resamples.}}
\label{fig:R1}
\vspace{3pt}
\end{figure}

\textcolor{red}{This obscuring effect reveals that the error formation process is more complex than the commonly assumed autoregressive iterative amplification. The propagation of existing errors by the ground-truth dynamics ($G$) can result in significant reduction, rather than merely amplification. As shown in Figure~\ref{fig:R1}b, while $G$ aligns with conventional understanding by dominating the $\|e_k\|^2$ at longer steps, its short-term behavior is quite complex and fluctuates in sign. As indicated by the green (FNO) and blue (UNet H1) lines, this negative contribution masks the severe local prediction error $J$ (Figure~\ref{fig:R1}c), which surprisingly results in these two solvers exhibiting the lowest $\|e_k\|^2$ (Figure~\ref{fig:R1}a). Furthermore, the coupling $C$ between the propagation and the local prediction error can also contribute negatively to $\|e_k\|^2$, which further explains the obscuring effect previously observed in UNet H1.}

\addtocounter{figure}{-1}

\subsection{Improved Forecast Accuracy \textcolor{red}{Often} Accompany Distorted Ensemble Evolution}

\textcolor{red}{While neural solvers are typically trained to minimize point-wise rollout errors, does this guarantee a physically consistent evolution of the entire state space? To answer this, we shift our focus from single-trajectory accuracy to the collective behavior of the ensemble.}

\textcolor{red}{
We use bold letters $\mathbf{x}_k, \hat{\mathbf{x}}_k$ to denote the ensemble states, with \(F\) acting memberwise.} Let \(\mathcal{Z}\) denote the centering operator that subtracts the ensemble mean, so that \textcolor{red}{$\mathcal{Z}\mathbf{x}_k$ represents the centered ensemble of $\mathbf{x}_k$. }

\textcolor{red}{The norm of the centered ensemble $\|\mathcal{Z}\mathbf{x}\|_M^2$ quantifies the size of the ensemble cloud,} where \(\|A\|_M^2 = \frac{1}{MD}\sum_{m=1}^M\sum_{i=1}^D A_{m,i}^2\) averages over members and spatial grid points. We then measure two cumulative effects of the local residual over a forecast window \(\mathcal{W}\).

\textcolor{red}{Define the \textit{relative ensemble size change} $D$ to quantify whether the solver's propagation compresses or expands the ensemble cloud compared to DNS propagation:
\begin{equation}
D = \sum_{k\in\mathcal{W}}\left(\|\mathcal{Z}\hat{\mathbf{x}}_{k+1}\|_M^2 - \|\mathcal{Z}F(\hat{\mathbf{x}}_k)\|_M^2\right).
\end{equation}}
A negative $D$ indicates that the solver relatively compresses the cloud, while a positive value indicates relative expansion.

\textcolor{red}{
Define the \textit{relative deviation from the true dynamics} $T$ to quantify whether the solver's propagation moves the ensemble cloud closer to or further from the ground truth compared to DNS propagation:
\begin{equation}
T = \sum_{k\in\mathcal{W}}\left(\|\mathcal{Z}\hat{\mathbf{x}}_{k+1} - \mathcal{Z}\mathbf{x}_{k+1}\|_M^2 - \|\mathcal{Z}F(\hat{\mathbf{x}}_k) - \mathcal{Z}\mathbf{x}_{k+1}\|_M^2\right).
\end{equation}}
A positive $T$ indicates that the solver pushes the cloud further from the ground truth ensemble \textbf{compared with true dynamics(DNS)}, whereas a negative value indicates it moves closer.

\begin{figure}[!htbp]
\centering
\includegraphics[width=\linewidth]{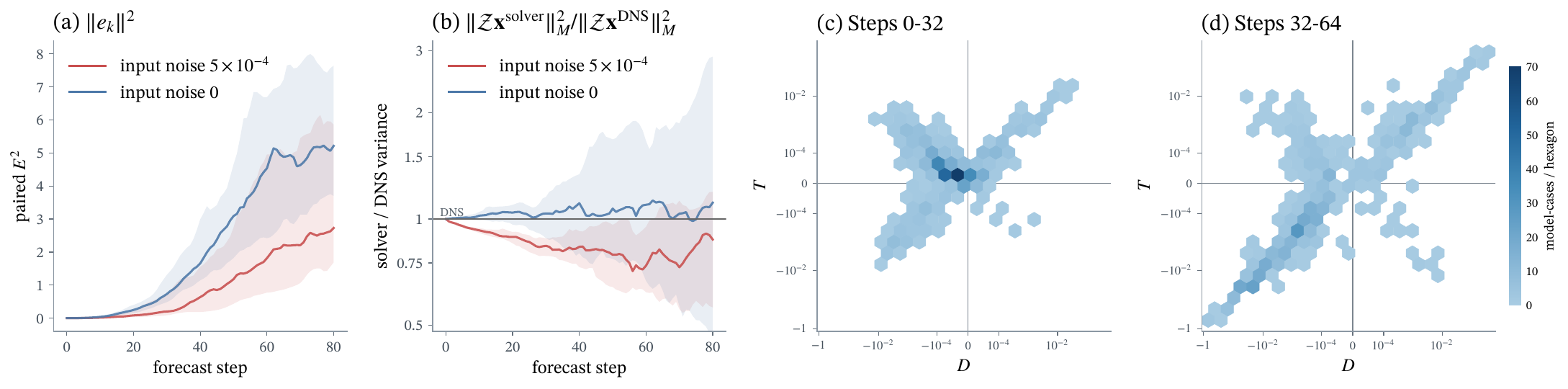}
\caption{\emph{Improved forecast accuracy often accompany ensemble contraction } \textbf{(a)} Rollout prediction error. \textbf{(b} Relative size of ensemble cloud. \textcolor{red}{In (a--b), lines and bands show the median and 25th--75th quantiles across cases.} \textbf{(c, d)} Joint distribution of the cumulative \(D\) (horizontal, relative ensemble size change) and  \(T\) (vertical, relative deviation from the true dynamics) across all tested solvers and cases. Each hexagon is colored by the number of (model, case) pairs it contains.}
\label{fig:dpot}
\vspace{3pt}
\end{figure}

Taking DPOT \citep{hao2024dpot} as an entry point, its training maps noisy states to clean next states; we construct a control by setting the input noise to zero. As shown in Figure~\ref{fig:dpot}a--b, this denoising procedure lowers the rollout prediction error but overly compresses the ensemble relative to the DNS reference. \textcolor{red}{Isolating this training design via our metrics, the first 32 steps exhibit \(D < 0\) and \(T > 0\), indicating that this compression fails to bring DPOT closer to the true dynamics and merely reduces the rollout error.}

Expanding this analysis to all tested solvers (Figure~\ref{fig:dpot}c-d), \textcolor{red}{we emphasize that \(D\) and \(T\) evaluate the solver's propagation from an existing predicted state relative to the DNS reference. Figure~\ref{fig:dpot}c illustrates the early rollout (step 0-32), where solvers predominantly cluster on the left (\(D < 0\)), revealing that ensemble contraction is a widespread phenomenon. As the rollout progresses (Figure~\ref{fig:dpot}d, step 32-64), the distribution concentrates along the bottom-left to top-right diagonal. This suggests that an ensemble approaching the true dynamics (smaller \(T\)) is consistently accompanied by stronger ensemble contraction (more negative \(D\)). Ideally, \(D\) should remain near zero to maintain a correct ensemble size; the observed trade-off highlights the inherent limitations of these solvers. Worse still, data points in the top-left region indicate a dual failure: the solvers overly compress the ensemble (\(D < 0\)) while simultaneously becoming less accurate (\(T > 0\)).}

\begin{findingbox}
\textbf{Main Findings}: Lower \textcolor{red}{rollout prediciton error} can obscure larger \textcolor{red}{local prediction error}. \textcolor{red}{The propagation of ground-truth dynamics can strongly contract existing errors.} Ensemble contraction is a widespread phenomenon. \textcolor{red}{Approaching the ground-truth ensemble is often accompanied by stronger ensemble contraction.}
\end{findingbox}

\section{Do Solvers Preserve the Geometry of Nearby Trajectories?}
\label{sec:geometry}
We examine the ensemble geometry: how widely trajectories spread, how many directions account for that spread, and whether those directions agree with the reference dynamics.

For each case, we center the vorticity fields over the ensemble, yielding the centered fields $z_{c,m} = x_{c,m} - \frac{1}{M}\sum_{n=1}^M x_{c,n}$. Let \(\|\cdot\|_{\text{RMS}}\) denote the spatial RMS norm. \textcolor{red}{Define \textit{ensemble spread} as $S_c(t) = \left[ \frac{1}{M} \sum_{m=1}^{M} \|z_{c,m}(t)\|_{\text{RMS}}^2 \right]^{1/2}$, indicating how widely the members spread.}

We perform proper orthogonal decomposition(POD) on \(z_{c,m}\), yielding orthogonal spatial modes \(u_{c,k}(t)\) and their corresponding projection coefficients. The variance captured by the \(k\)-th mode is its eigenvalue \(\lambda_{c,k}(t)\). \textcolor{red}{Define \textit{effective rank} as the minimum number of leading modes required to capture at least 90\% of the total variance: $rank_{c}(t) = \min \left\{ rank : \frac{\sum_{k=1}^{rank} \lambda_{c,k}(t)}{\sum_k \lambda_{c,k}(t)} \ge 0.9 \right\}$.} It quantifies the effective dimensionality of the ensemble.

Let \(U_c(t)\) be the matrix whose columns are the leading modes up to the effective dimension \(rank_c(t)\). \textcolor{red}{Define \textit {subspace reorientation} \(R_c(t)\) over a 10-step interval as: $R_c(t) = 1 - \frac{1}{rank_c(t)} \left\| U_c(t)^\top U_c(t+10) \right\|_F^2.$} \(R_c(t)\) quantifies the temporal change of the dominant subspace. \(R = 0\) indicates completely identical subspaces, while \(R = 1\) indicates orthogonal subspaces.

\begin{figure}[!htbp]
\centering
\includegraphics[width=\linewidth]{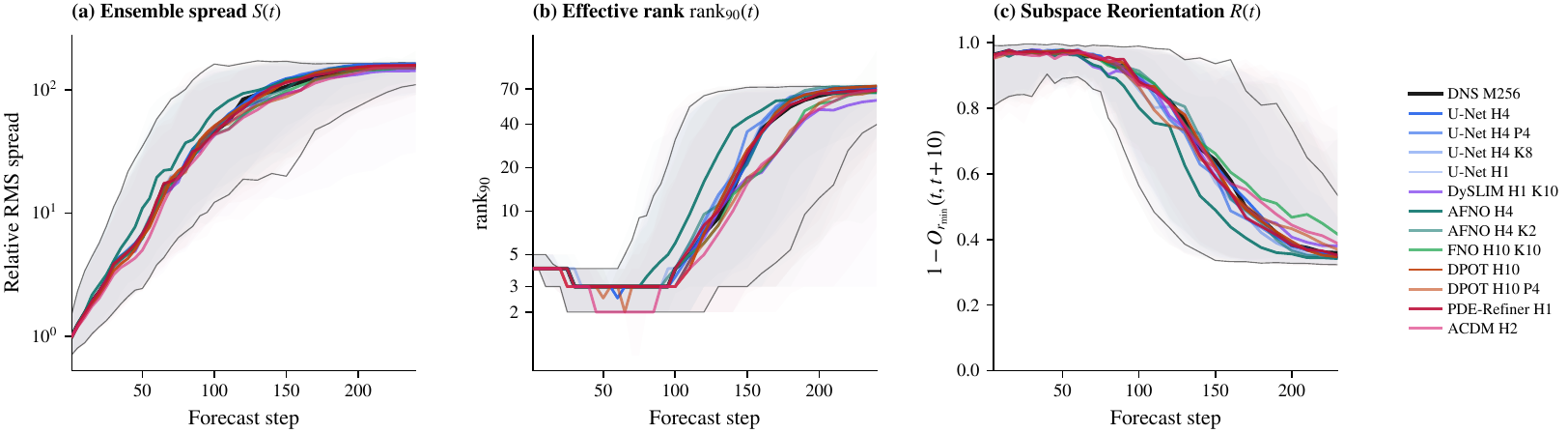}
\caption{\emph{Ensemble spread, effective rank, and subspace orientation evolve over time.} Solid curves show the median across all cases \(c\) at each time step. Shaded regions denote the 10th--90th quantile range, and the DNS quantiles are marked with thin black lines.}
\label{fig:geometry}
\vspace{3pt}
\end{figure}

Most tested neural solvers reproduce the broad trends in ensemble spread, effective dimension and temporal reorientation of reference dynamics. As trajectories diverge over time (Figure~\ref{fig:geometry}a), their variance initially remains confined to a few dominant directions before spreading across many more modes at later steps (Figure~\ref{fig:geometry}b). Concurrently, the dominant subspaces undergo continuous rotation (Figure~\ref{fig:geometry}c).

\begin{figure}[!htbp]
\centering
\includegraphics[width=\linewidth]{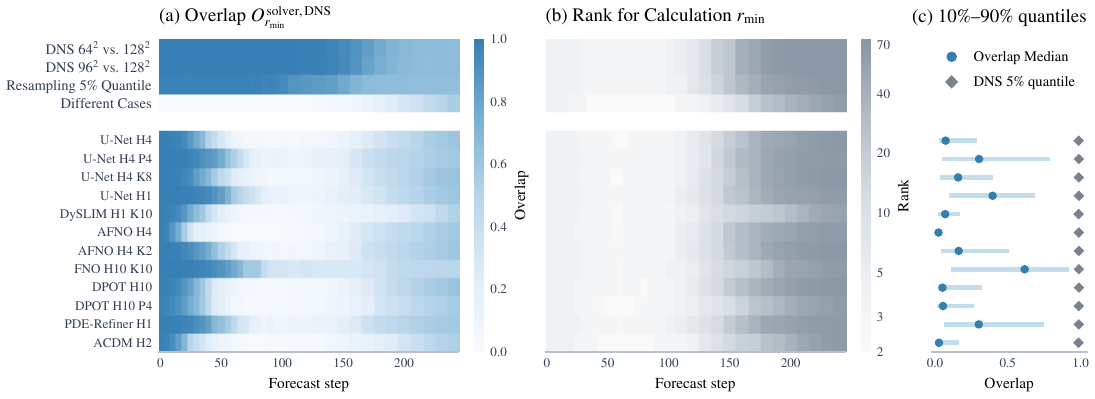}
\caption{\textcolor{red}{\emph{Similar ensemble spread and dimension fail to reveal misaligned dominant directions.} Rows 1--2 compare against coarse DNS, row 3 indicates the 5\% lower bound from resampling 256-member sets from a 1024-member DNS ensemble 100 times, and row 4 shows overlaps between different cases.} \textbf{(a)} $O(t)$ across test cases. \textbf{(b)} $r_{\min}(t)$ across test cases. \textbf{(c)} Overlap summarized by its temporal median over steps 50–90 for each case: dots show medians across cases, and bars show the 10th–90th percentiles. \textcolor{red}{Gray diamonds are the across-case median of per-case 5th quantiles from 100 rank-matched DNS resamples, not confidence bounds for the blue points.}}
\label{fig:overlap}
\vspace{3pt}
\end{figure}

\textcolor{red}{However, it remains unclear whether the specific orientations of these subspaces align with those of DNS.} To investigate this, we examine DNS-neural solver subspace overlap (Figure~\ref{fig:overlap}).

\textcolor{red}{Let \(O_{r_{\min}}^{\mathrm{solver,DNS}}(s)\) denote the \textit{subspace overlap} between the solver and DNS:}
\begin{equation}
    O_{r_{\min}}^{\mathrm{solver,DNS}}(t) = \frac{1}{r_{\min}} \left\| U_{r_{\min}}^{\mathrm{solver}}(t)^\top U_{r_{\min}}^{\mathrm{DNS}}(t) \right\|_F^2,
\end{equation}
where \(r_{\min} = \min(r_{90}^{\mathrm{solver}}, r_{90}^{\mathrm{DNS}})\). \textcolor{red}{Truncating both subspaces to the smaller effective rank \(r_{\min}\) ensures they share a common dimension for direct projection.}

\textcolor{red}{The low subspace overlap $O_{r_{\min}}^{\mathrm{solver,DNS}}$ directly answers our question: the solvers fail to capture the correct dominant directions, even when their ensemble spread and effective dimensions closely match the reference (Figure~\ref{fig:overlap}). Critically, this misalignment is far below both the coarse DNS baseline and the 5\% resampling lower bound, confirming it is a genuine failure of the neural solvers rather than an artifact of limited DNS resolution or finite ensemble size.} This misalignment occurs precisely while the ensemble variation remains concentrated in a few modes. Consequently, the trajectories diverge along fundamentally incorrect dominant spatial patterns. \textcolor{red}{Furthermore, the substantial variation across different initial conditions indicates that this subspace orientation captures highly informative and case-specific dynamics, rather than random noise.}

\begin{findingbox}
\textbf{Main Findings: Matching ensemble spread and effective dimension does not establish faithful perturbation dynamics.} Most tested solvers reproduce their broad temporal evolution but place the dominant variation along spatial directions that differ substantially from those of the reference dynamics.
\end{findingbox}

\section{Do Solvers Reproduce Extreme Events Under the Right Conditions?}
\label{sec:events}

We next examine whether neural solvers reproduce occurrence and evolution of intense vortices. Matching overall event frequency does not establish that events arise under the correct initial conditions, since \emph{overprediction} and \emph{underprediction} can cancel across cases and times. We first quantify these discrepancies, then examine whether predicted vortices deform and rotate consistently with DNS.

\subsection{Correct Aggregate Frequencies Can Conceal Incorrect Event Conditions}

We identify an event when a same-sign connected region with $|\omega|\geq10$ occupies at least 0.5\% of the periodic domain, and estimate its probability for each case and time as the fraction of ensemble members satisfying this criterion. Excess and Deficit separately average overprediction and underprediction relative to DNS, exposing errors that cancel in their signed difference, Bias. We weight cases equally and use trapezoidal time weights; all quantities are normalized by the mean DNS event frequency and reported as percentages (Appendix~\ref{app:event_definition}).

\begin{table}[!htbp]
\centering
\setlength{\abovecaptionskip}{0pt}
\setlength{\belowcaptionskip}{10pt}
\caption{Event-frequency errors across test cases, expressed as percentages of the mean reference event frequency. For stochastic solvers, S and I denote noise shared across or sampled independently for ensemble members.}
\label{tab:event_frequency}
\renewcommand{\arraystretch}{1.25}
\scriptsize 
\setlength{\tabcolsep}{1pt} 
\begin{tabular*}{\linewidth}{@{\extracolsep{\fill}}l c@{\hspace{1pt}}c@{\hspace{-2pt}} *{12}{r}@{}} 
\noalign{\hrule height 0.8pt}
 & \multicolumn{2}{c}{\textcolor{red}{Coarse DNS}} & \multicolumn{4}{c}{U-Net} & \multicolumn{2}{c}{AFNO} & \multicolumn{2}{c}{DPOT} & \multicolumn{2}{c}{PDE-Refiner} & ACDM & DySLIM \\
\cline{2-3}\cline{4-7}\cline{8-9}\cline{10-11}\cline{12-13}\cline{14-14}\cline{15-15}
 & \textcolor{red}{64\(^2\)} & \textcolor{red}{96\(^2\)} & H1 & H4 & H4 P4 & H4 K8 & H4 & H4 K2 & H10 & H10 P4 & H1 S & H1 I & H2 S & H1 K10 \\
\hline
Bias (\%) & \textcolor{red}{-0.01} & \textcolor{red}{-0.48} & +0.42 & -60.36 & +60.83 & -44.33 & -6.93 & -16.51 & -38.43 & -25.85 & +0.44 & +4.74 & -22.27 & +17.16 \\
Excess (\%) & \textcolor{red}{7.35} & \textcolor{red}{6.06} & 41.31 & 11.08 & 81.17 & 17.06 & 53.42 & 35.57 & 23.39 & 27.16 & 36.74 & 41.38 & 38.86 & 54.29 \\
Deficit (\%) & \textcolor{red}{7.35} & \textcolor{red}{6.54} & 40.89 & 71.44 & 20.34 & 61.39 & 60.35 & 52.08 & 61.83 & 53.01 & 36.30 & 36.64 & 61.13 & 37.13 \\
\noalign{\hrule height 0.8pt}
\end{tabular*}
\vspace{3pt}
\end{table}

As can be observed in Table~\ref{tab:event_frequency}, substantial overprediction and underprediction occur across the tested solvers, including those with nearly correct aggregate frequencies. U-Net H1, for example, has a normalized bias of only $0.42\%$, yet its excess and deficit are $41.31\%$ and $40.89\%$, respectively. Similarly, PDE-Refiner with shared noise has a bias of $0.44\%$ but a combined absolute error of $73.04\%$. Their close aggregate agreement results from cancellation across initial-state ensembles and times, rather than consistently accurate event probabilities.

\begin{figure}[!htbp]
\centering
\includegraphics[width=\linewidth]{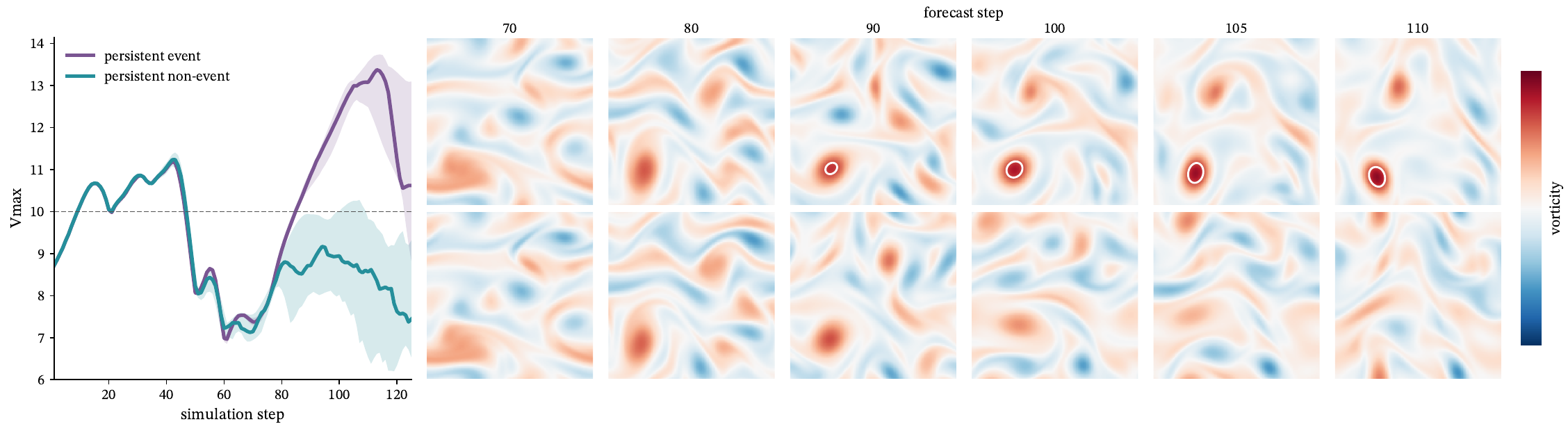}
\caption{Nearby initial states lead to distinct persistent event outcomes in DNS. For case 1616, peak-vorticity histories and representative fields are shown for groups defined by persistent event and non-event outcomes.}
\label{fig:dns_groups}
\vspace{3pt}
\end{figure}

\begin{table}[!htbp]
\centering
\setlength{\abovecaptionskip}{0pt}
\setlength{\belowcaptionskip}{10pt}
\caption{Predicted event frequency (\%) within the fixed DNS groups of case 1616, averaged over members and the evaluation window defined in Appendix~\ref{app:event_groups}. DNS frequencies are given in parentheses.}
\label{tab:branch_frequency}
\footnotesize
\renewcommand{\arraystretch}{1.25}
\setlength{\tabcolsep}{1.5pt}
\begin{tabular*}{\linewidth}{@{\extracolsep{\fill}}l*{11}{r}@{}}
\noalign{\hrule height 0.8pt}
 & \multicolumn{4}{c}{U-Net} & \multicolumn{2}{c}{AFNO} & \multicolumn{2}{c}{DPOT} & \mbox{PDE-Refiner} & ACDM & DySLIM \\
\cline{2-5}\cline{6-7}\cline{8-9}\cline{10-10}\cline{11-11}\cline{12-12}
 & H1 & H4 & H4 P4 & H4 K8 & H4 & H4 K2 & H10 & H10 P4 & H1 S & H2 S & H1 K10 \\
\hline
Event (100\%) & 0.00 & 0.00 & 100.00 & 0.00 & 0.00 & 0.00 & 0.00 & 0.00 & 0.00 & 0.00 & 0.00 \\
Non-event (0\%) & 0.00 & 0.00 & 69.58 & 0.00 & 0.00 & 6.99 & 0.00 & 0.00 & 0.00 & 4.90 & 0.00 \\
\noalign{\hrule height 0.8pt}
\end{tabular*}
\vspace{3pt}
\end{table}

For hazardous events, false negatives can be more consequential than false positives because they leave the hazard unanticipated. We therefore examine whether solvers recover extreme events that persist under the reference dynamics in case 1616 (Figure~\ref{fig:dns_groups}). We fix event and non-event groups using the DNS outcomes and evaluate each solver on the same members (Appendix~\ref{app:event_groups}). Ten of the eleven compared configurations predict no events in the DNS event group (Table~\ref{tab:branch_frequency}). The remaining configuration, U-Net H4 P4, recovers all events in that group but also predicts events with a frequency of $69.58\%$ in the DNS non-event group. Thus, the dominant failure in this case is the complete omission of persistent reference events; the only configuration that recovers them also produces frequent false positives.

\subsection{Recognizable Vortices Can Deform and Rotate Incorrectly}

Event occurrence captures only part of the dynamics. Once a vortex forms, its deformation and rotation reflect its interaction with the surrounding flow. In the classical elliptic-vortex problem, background strain and vorticity govern changes in axis ratio and orientation \citep{kida1981motion}. These geometric properties therefore provide diagnostics beyond peak intensity.

In case 4040, we characterize the evolution of the dominant vortex core using its peak vorticity $V$, the major-to-minor axis ratio $\chi_{70}$ of its 70\%-of-peak core, and the accumulated change in its major-axis orientation (Appendix~\ref{app:vortex_geometry}).

\begin{figure}[!t]
\centering
\includegraphics[width=\linewidth]{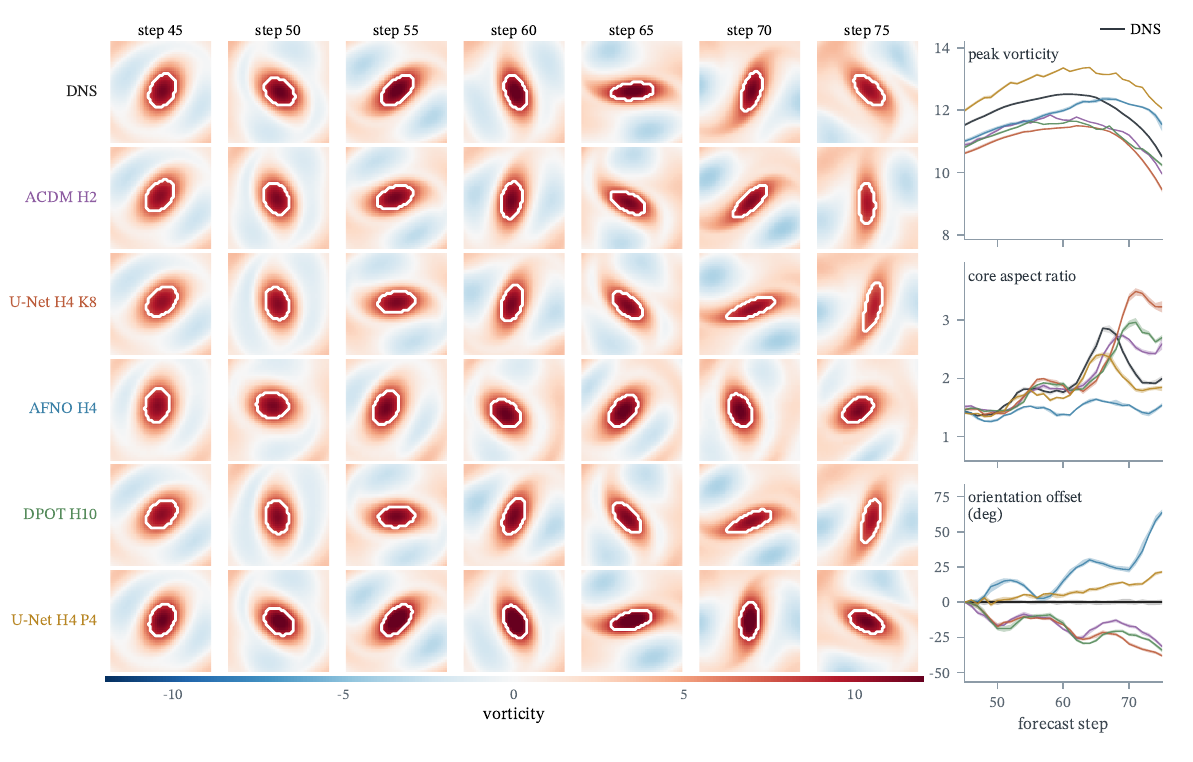}
\caption{Predicted vortices can retain similar intensity while deforming and rotating differently from DNS. Case 4040. Left: fields evolved from the same initial perturbation, with each solver's 70\%-of-peak core boundary shown in white. Right: medians across members for peak vorticity $V$, core axis ratio $\chi_{70}$, and accumulated major-axis rotation relative to the DNS median. \textcolor{red}{Bands show the 10th--90th member quantiles.}}
\label{fig:local_dynamics}
\vspace{3pt}
\end{figure}

Figure~\ref{fig:local_dynamics} shows that the solvers preserve recognizable intense vortices but misrepresent their subsequent geometry. AFNO retains a comparatively round core while the DNS vortex elongates. Conversely, U-Net H4 K8 and DPOT sustain elongation after the reference vortex becomes rounder. Predicted orientations also advance ahead of or lag behind DNS. These discrepancies persist even when peak intensities are similar, showing that a coherent appearance and plausible strength do not establish correct local evolution.

\begin{findingbox}
\textbf{Main Findings: Near-correct aggregate event frequencies can hide substantial errors across initial conditions and times.} In the examined branch-separation case, 10/11 configurations miss the persistent event group entirely, while the remaining configuration also predicts frequent events in the non-event group. Even when vortices remain recognizable and have similar peak intensities, their deformation and rotation can differ substantially from DNS.
\end{findingbox}

\section{Implications for Evaluating Neural PDE Solvers}
\label{sec:implications}

Our experiments reveal three gaps between predictive performance and learned dynamics: (1) lower trajectory error can coexist with inaccurate local updates, (2) similar ensemble spread can hide incorrect directions of evolution, and (3) matching event frequencies can conceal missed event-producing conditions. Although these gaps are only demonstrated on turbulent Kolmogorov flow (see Appendix~\ref{app:discussion} for limitations and future directions), these findings motivate a broader principle: \emph{\textbf{training and evaluation should account for the physical dynamics that matter to the intended application}}.

\noindent\textbf{Assess how training objectives affect dynamics.}
In a chaotic system, nearby states follow distinct trajectories. Training perturbed and unperturbed states toward the same next-frame target, as in denoising, encourages insensitivity that may remove observation noise but may also suppress physically meaningful differences. Multi-step rollout training presents a related tension: states displaced by earlier prediction errors are trained toward the original reference trajectory, whose continuation may differ from their own. Perturbation scales, forecast intervals, and unroll lengths should therefore be evaluated for their effects on both prediction accuracy and the state evolution during rollout.

\noindent\textbf{Design benchmarks around physical systems and their uses.}
Datasets and benchmark platforms such as PDEBench \citep{takamoto2022pdebench}, The Well \citep{ohana2024well}, BLASTNet \citep{chung2023turbulence}, and PDEArena \citep{gupta2022towards} broaden access to physical simulation data and tools for evaluating neural solvers. To make comparisons scientifically informative, benchmarks should specify whether the dynamics are deterministic or stochastic, whether the observed state is sufficient to determine its subsequent evolution, and which time scales and physical outcomes matter. These considerations should guide the inputs, forecast horizons, and evaluation criteria. For hazard prediction, for example, identifying event-producing conditions may be more important than a modest reduction in average field error. Broad coverage should therefore be accompanied by diagnostics tailored to each system and task.

\noindent\textbf{Make dynamical fidelity part of the evaluation standard.}
Evaluation should establish whether gains in prediction scores also improve relevant dynamical behavior. Our investigation provides three complementary checks: agreement of local updates with reference evolution, preservation of ensemble geometry, and reproduction of extreme events under the appropriate conditions. Other systems may require different diagnostics, but the aim is the same: identify which physical behaviors improve and which remain distorted. For stochastic solvers, predictive-distribution quality and the dynamical fidelity of generated trajectories require separate assessment. Reporting dynamical diagnostics alongside conventional scores like prediction error would expose otherwise hidden trade-offs and provide stronger evidence of progress in neural PDE modeling.

\clearpage
\subsection*{AI use statement}
We used generative AI tools for language polishing, assistance with code development, and checking for potentially overlooked related work. The authors reviewed and revised AI-assisted text and inspected and tested AI-assisted code before use. All suggestions were evaluated against the intended methods and claims. The authors take full responsibility for the manuscript, code, experimental results, and conclusions.

\bibliography{references}
\bibliographystyle{references_style}

\clearpage
\appendix
\addtocontents{toc}{\protect\setcounter{tocdepth}{2}}
\begingroup
\renewcommand{\contentsname}{Appendix Contents}
\tableofcontents
\endgroup
\clearpage

\section{Discussion and Future Work}
\label{app:discussion}

Our experiments are limited to two-dimensional Kolmogorov flow. This spatially periodic shear flow provides a simple setting for studying hydrodynamic instability \citep{meshalkin1961investigation} and the recurrent flow structures underlying turbulent motion \citep{chandler2013invariant}. Its tractable formulation makes it useful for examining learned dynamics, but results from one system do not establish how broadly the observed failures extend. Tests on additional dynamical systems may yield a more comprehensive understanding and reveal different relationships between prediction accuracy and dynamical fidelity.

Our diagnostics identify how predicted dynamics depart from DNS, but do not establish which architectural or training choices cause these departures. Prior studies show that both model design and training objectives can affect rollout behavior \citep{mccabe2023towards,lippe2023pde}. Establishing which factors account for the failures observed here requires further controlled comparisons. More broadly, future work should examine whether improvements in dynamical fidelity persist across systems and operating conditions, alongside gains in conventional predictive performance.

\section{Related Work}
\label{app:related}

\subsection{Ensemble Forecasting}
\label{app:ensemble-literature}

\noindent\textbf{Forecasting with perturbed initial states.}
\citet{leith1974theoretical} studied finite Monte Carlo ensembles in a test-field model of two-dimensional turbulence to assess forecast skill when the atmosphere is incompletely observed. Ensemble-based estimates improved mean-square wind forecasts, in part by filtering unpredictable small-scale structure; the useful forecast range depended on observational resolution. In operational weather prediction, \citet{toth1993ensemble} used ensembles at NMC to represent growing analysis errors and obtain useful forecasts with few members. Their preliminary experiments found that two bred-perturbation forecasts could outperform substantially larger random or lagged-forecast ensembles. Ensembles also support investigations of dynamical sensitivity: \citet{vela2024large} used DNS ensembles with nearby initial states in Kolmogorov flow to study the predictability of extreme dissipation bursts, connecting event predictability to large-scale circulation patterns.

\noindent\textbf{What the perturbed states represent.}
In operational forecasting, perturbed states primarily represent uncertainty about the actual initial state. This uncertainty arises from measurement errors, incomplete observational coverage, and errors in the background forecast used in data assimilation \citep{leutbecher2008ensemble}. Perturbations therefore represent plausible errors of the estimated state, rather than measurement noise alone. Forecast-model errors can also affect that estimate through assimilation; unresolved processes, uncertain parameterizations, and numerical approximations contribute further uncertainty during forecast evolution. Ensemble-based assimilation seeks to sample the initial uncertainty distribution, while dynamically constrained approaches select forecast-sensitive directions when that distribution cannot be reliably estimated \citep{leutbecher2008ensemble}.

\noindent\textbf{Perturbed states as dense probes of learned dynamics.}
Many neural PDE solvers are trained and evaluated on numerical simulation data, as in PDEBench and The Well \citep{takamoto2022pdebench,ohana2024well}. When complete simulated states are supplied without observation noise, incomplete observations and measurement errors are absent from the learning problem. A numerical simulator may differ from the physical system it approximates, but that discrepancy does not itself make the simulator's specified initial state or deterministic evolution uncertain. Errors in learning that evolution are a separate issue. Our motivation is to probe this learned evolution: we deliberately construct dense ensembles of nearby initial states and compare their evolution under each solver and the reference dynamics. \emph{These perturbations provide controlled probes of local dynamical behavior, rather than estimates of uncertainty about an unknown initial state.} Their evolving errors, geometry, and extreme-event behavior reveal how the learned dynamics respond to nearby states.

\subsection{Evaluation of Neural PDE Solvers}
\label{app:solver-evaluation-literature}

\noindent\textbf{Large datasets and benchmarks.}
Evaluation suites examine neural solvers across equations, parameters, and prediction horizons. PDEBench includes field errors, errors in conserved quantities and boundary values, and errors separated into Fourier bands \citep{takamoto2022pdebench}. PDEArena compares neural architectures and their generalization across PDE parameters and time scales \citep{gupta2022towards}. The Well reports variance-normalized errors for individual steps and rollout windows, and provides spectral diagnostics to resolve performance across spatial scales \citep{ohana2024well}. BLASTNet complements forecasting benchmarks with three-dimensional turbulence data and a volumetric super-resolution study \citep{chung2023turbulence}. APEBench provides a configurable benchmark across PDE dynamics, architectures, and unrolled training procedures, including experiments connecting rollout accuracy to the learned numerical update \citep{koehler2024apebench}.

\noindent\textbf{Long-term statistics and dynamical invariants.}
Learned models have also been evaluated beyond trajectory agreement. \citet{park2024dynamical} provide an ergodic-theoretic account of why small prediction errors can coexist with inaccurate invariant statistics, and study how incorporating Jacobian information during training improves statistical accuracy. \citet{pathak2017using} used reservoir computing to reproduce the attractor statistics and estimate Lyapunov exponents of the Lorenz system and the Kuramoto--Sivashinsky equation. \citet{jiang2023training} studied neural operators trained on noisy Lorenz-96 and Kuramoto--Sivashinsky data using distributional and contrastive objectives. Their evaluation includes summary-statistic distributions and energy spectra, together with Lyapunov and attractor-dimension diagnostics for Lorenz-96; the results show improved long-term behavior relative to training only on prediction error. DySLIM evaluates learned chaotic dynamics using distributional distances, energy spectra, spatial covariance, and temporal correlations \citep{schiff2024dyslim}. Its invariant-measure regularization improves rollout stability and statistical agreement in the tested systems. For stochastic PDEs, \citet{srikishan2026trie} introduce TRIE, which evaluates invariant-measure fidelity, predictive uncertainty calibration, and computational efficiency on stochastic Kuramoto--Sivashinsky and Kolmogorov systems. These studies evaluate learned dynamics through long-term statistics, dynamical invariants, and predictive distributions.

{\color{red}
\noindent\textbf{Error growth, propagation, and reference-solver interventions.}
Several neural-PDE studies characterize how errors grow over autoregressive rollouts. PDE-Refiner describes the propagation of rollout error and, for the Kuramoto--Sivashinsky equation, how nonlinear interactions transfer high-frequency errors into lower frequencies \citep{lippe2023pde}. \citet{mccabe2023towards} analyze uncontrolled autoregressive error growth and eventual instability, while DySLIM relates the instability of learned chaotic dynamics to positive Lyapunov exponents that exponentially amplify errors \citep{schiff2024dyslim}. These works primarily characterize rollout-level amplification, spectral contamination, and instability.

More recent analyses explicitly separate one-step defects from their subsequent propagation. \citet{ainslie2026eigenanalysisframeworkautoregressiveneural} linearize the learned one-step map along the true trajectory, decomposing error into an injected one-step term and a term propagated by the learned-map Jacobian. \citet{bangera2026doesjacobianfinetuningstabilize} additionally construct a counterfactual recursion that propagates the same measured one-step defects through the true solver Jacobian, thereby separating physical instability from excess amplification introduced by the emulator. Our decomposition differs by directly evaluating the full nonlinear reference evolution from the predicted state, $F(\hat{x}_k)-F(x_k)$, rather than approximating this propagation with a Jacobian along the true trajectory. It therefore gives an exact finite-error identity at each step and in cumulative form, and its signed squared-error contributions expose both amplification and strong contraction by the reference dynamics.

Reference solvers have also been applied directly to predicted states for training or correction. APEBench introduces diverted-chain training, in which a differentiable reference solver advances each autoregressively predicted state by one step to provide a training target \citep{koehler2024apebench}. ANCHOR adaptively invokes a high-fidelity numerical solver during neural-operator rollouts and reports that these interventions rectify accumulated errors \citep{roy2026anchorerrorcontrolledadaptivenumerical}. These methods use the reference solver to improve a model or its rollout. In contrast, we use matched DNS advancements diagnostically to isolate and quantify how local prediction errors and reference-dynamics propagation jointly form the observed rollout error.
}

\noindent\textbf{Local dynamical properties of forecasts.}
\citet{margazoglou2023stability} evaluate whether echo state networks recover tangent-space dynamics, including covariant Lyapunov vectors, finite-time Lyapunov exponents, and angles between stable, neutral, and unstable subspaces. \citet{fang2025dynamical} directly assess dynamical discrepancies in machine-learning forecasts of Lorenz, Kuramoto--Sivashinsky, and Kolmogorov systems. They estimate instantaneous dimension and inverse persistence from recurrences in a training-data reference attractor, relate these quantities to conventional forecast errors, and measure errors in the dynamical indices themselves. States with higher dimension and lower persistence tend to have larger forecast errors, while recursive forecasts can progressively distort these local properties. These approaches characterize tangent-space stability and local attractor properties. Our ensemble diagnostics instead follow finite initial-state perturbations and compare their evolving spread and dominant spatial directions with those of the reference dynamics.

\section{DNS, Dataset Construction, and Forecast Protocol}
\label{app:setup}

\noindent\textbf{DNS Settings.} DNS uses a $128{\times}128$ Fourier pseudospectral discretization on $[0,2\pi)^2$, with viscosity $\nu=0.01$ and an internal time step of $0.002$. The spectral update applies the two-thirds de-aliasing rule, removing modes with $|k_x|>42$ or $|k_y|>42$. Vorticity fields are saved at intervals of $\Delta t=0.2$, giving $D=128^2$ spatial degrees of freedom. In this two-dimensional incompressible periodic flow, the spatial mean velocity is fixed at zero, and the velocity field is recovered from vorticity by solving the streamfunction Poisson equation \citep{vela2024large}. With the forcing and viscosity fixed, the full vorticity field therefore specifies the dynamical state and determines its subsequent reference evolution.

\noindent\textbf{Dataset and Perturbations.} The archive provided by \citet{vela2024large} contains $8,192$ temporally decorrelated states of statistically stationary Kolmogorov flow. We sample 512 base initial states, each defining a case $c$. Around each base state, we generate spatially structured perturbations from randomly signed differences between sampled archive states, rescaled to a spatial mean square of $10^{-5}$. Each perturbed initial state defines an ensemble member $m$ and is shared across DNS and the neural solvers; member 0 denotes the unperturbed trajectory.

The perturbation generator forms $\delta\omega=s(\omega_a-\omega_b)$, where $s\in\{-1,+1\}$ and the two distinct archive states exclude the base state. It rescales each difference by $\sqrt{10^{-5}/\|\delta\omega\|_{\mathrm{sp}}^2}$ before adding it to the base state, so the prescribed scale is a spatial mean square, not an RMS amplitude. The generation configuration uses balanced archive-state pairing and signs; ensemble diagnostics use the matched subset of 256 perturbed members rather than all generated members.

We divide the cases into 416 for training, 48 for validation, and 48 for testing. Training and validation use only unperturbed trajectories. Dynamical evaluation uses $M=256$ perturbed members per test case, with a complete DNS reference trajectory for every member. The unperturbed member is excluded from ensemble statistics and evaluated separately using conventional forecast metrics.

\noindent\textbf{Forecast Origin and Stochastic Sampling.} Perturbed initial states first evolve under DNS for ten saved time steps to provide input histories. Each solver receives its required input history from data steps 1--10 and predicts from data step 11 onward, using its own outputs for subsequent updates. Predictions are compared with DNS at matching times. Throughout the paper, prediction step 1 corresponds to data step 11. For stochastic solvers, all members within a case share a random path, with fresh draws at successive updates and different paths across cases, so that sampling differences do not confound responses to initial-state perturbations. We generate one predicted trajectory per initial state; experiments using independent randomness across members are identified separately.

{\color{red}
\noindent\textbf{Coarse DNS Resolution Controls.}
We run DNS ensembles at spatial resolutions $N\times N$
with $N\in\{64,96,128\}$.
All three resolutions use the same governing equations and
numerical rules on the periodic domain $[0,2\pi)^2$, with
viscosity $\nu=0.01$, vorticity forcing $-2\sin(4y)$,
an internal time step of $0.002$, and an output interval
of $0.2$.
Strict two-thirds de-aliasing retains only Fourier modes
satisfying $3|k_x|<N$ and $3|k_y|<N$.
The largest retained wavenumber in each coordinate is
therefore 21, 31, and 42 for $N=64$, 96, and 128,
respectively.

For each ensemble member, all three resolutions start from
the same state at the common forecast origin
(absolute data step 10).
We spectrally project this state onto each resolution
without renormalizing the perturbation after projection.
Each resolution is then evolved independently.
The coarse trajectories are therefore not downsampled
versions of the $128^2$ trajectory: removing high-wavenumber
modes also changes subsequent nonlinear interactions.

For comparison, we lift the $64^2$ and $96^2$ outputs to
the $128^2$ grid by zero-padding their Fourier
representations, while retaining all resolved modes of
the $128^2$ output.
The $128^2$ run serves as the reference for this
resolution control; convergence to the continuum limit
has not been established.}

\section{Solver Configurations and Forecast Evaluation}
\label{app:models}
We screen configurations using conventional relative $L_2$ forecast error, excluding PFNO H10 P10, DYffusion H1 P16, FNO H1, and DPOT H10 trained from scratch and retaining 12 configurations for dynamical evaluation. Table~\ref{tab:training} records the training configurations; Table~\ref{tab:models} reports their parameter counts and forecast errors. Scores are case medians of relative $L_2$ field error on the unperturbed test trajectories, using the common forecast origin. Stochastic solvers use one fixed noise realization per case. Training objectives differ across configurations; controlled interventions hold the remaining training settings fixed.

\paragraph{Architectures and learning procedures.}
U-Net uses a convolutional encoder--decoder with skip connections between corresponding resolutions \citep{ronneberger2015u}. FNO learns convolution kernels through Fourier-mode weights \citep{li2020fourier}; AFNO uses structured channel mixing in Fourier space \citep{guibas2021adaptive}. DPOT combines a Fourier-based operator architecture with autoregressive denoising pretraining \citep{hao2024dpot}. Our denoising control is described in the main text. DySLIM is a training framework that regularizes learned dynamics through distribution matching, rather than a distinct network architecture \citep{schiff2024dyslim}.

U-Net has base width 64, channel multipliers $(1,2,4,4)$, two residual blocks per level, and circular padding. AFNO uses $8\times8$ patches, embedding width 384, eight layers, eight Fourier channel blocks, an MLP expansion factor of four, and a soft-threshold parameter of 0.01. FNO H10 K10 uses width 32, five layers, 32 retained modes in each spatial direction, and positional grid inputs. DPOT uses the tiny configuration with $8\times8$ patches, width 512, four layers, four blocks, and 32 modes; the operator blocks, positional embeddings, and temporal aggregation weights are initialized from the pretrained checkpoint.

The U-Net H4 K8 configuration starts from the H4 checkpoint and advances seven updates without gradients before differentiating the eighth-step loss. AFNO H4 K2 starts from its one-step H4 checkpoint and differentiates through both updates. AFNO and FNO use relative $L_2$ loss; their multi-step configurations sum the equally weighted losses without dividing by the unroll length. These settings distinguish pushforward training from full differentiation through a rollout.

DySLIM uses a periodic convolutional network of width 200, with four levels, four processors, $3\times3$ kernels, and an input residual connection. Its loss combines squared Sobolev $H^1$ error with conditional MMD. MMD uses a biased multiscale inverse-multiquadric estimator on a stride-two view of the fields. The rollout curriculum samples lengths up to ten, with gradients stopped through preceding updates. PDE-Refiner uses a conditioned U-Net of base width 32, channel multipliers $(1,1,2,4)$, two residual blocks per level, and circular padding; it predicts state differences.

\paragraph{Stochastic forecasting models.}
PDE-Refiner uses successive denoising refinements to recover components that can be underrepresented by a conventional prediction loss \citep{lippe2023pde}. ACDM generates the next flow state by conditional diffusion given preceding states \citep{kohl2026benchmarking}. DYffusion uses learned temporal interpolation within a diffusion-based forecasting procedure \citep{ruhling2023dyffusion}; the evaluated configuration is included below among the excluded settings. Their original evaluations address different targets: PDE-Refiner emphasizes accurate rollout duration, while ACDM examines flow statistics and stability, and DYffusion reports probabilistic forecast scores. Here, their sampled trajectories follow the common initialization and randomness protocol in Appendix~\ref{app:setup}. These descriptions distinguish the published methods from the particular configurations evaluated in this study.

\begin{table}[!htbp]
\centering
\caption{Training configurations. $H/P/K$ gives history, prediction block, and training unroll length. Validation losses select checkpoints within each training setup and are not comparable forecast scores across objectives.}
\label{tab:training}
\small
\begin{tabular}{lp{0.13\linewidth}p{0.49\linewidth}}
\hline
Configuration & $H/P/K$ & Training objective and procedure\\
\hline
U-Net H1 & 1/1/1 & MSE, one-step training\\
U-Net H4 & 4/1/1 & MSE, one-step training\\
U-Net H4 P4 & 4/4/1 & MSE, direct four-step targets\\
U-Net H4 K8 & 4/1/8 & MSE, eight-step pushforward training\\
AFNO H4 & 4/1/1 & Relative $L_2$, one-step training\\
AFNO H4 K2 & 4/1/2 & Relative $L_2$, two-step full unroll\\
FNO H10 K10 & 10/1/10 & Relative $L_2$, summed over ten unrolled steps\\
DPOT H10 & 10/1/1 & Fine-tuned from the released pretrained checkpoint\\
DPOT H10 P4 & 10/4/1 & Fine-tuned from the released pretrained checkpoint\\
DySLIM H1 K10 & 1/1/10 & Sobolev $H^1$ plus conditional MMD; pushforward curriculum\\
PDE-Refiner H1 & 1/1/1 & Diffusion denoising MSE\\
ACDM H2 & 2/1/1 & DDPM denoising MSE; 20 sampling steps\\
\hline
\end{tabular}
\end{table}

\begin{table}[!htbp]
\centering
\caption{Solver configurations and conventional forecast accuracy. Columns report parameter counts and case-median relative $L_2$ field error, $\|\hat x-x\|_2/\|x\|_2$, on test trajectories.}
\label{tab:models}
\small
\begin{tabular}{lrrr}
\hline
Configuration & Parameters (M) & Step 20 & Step 80\\
\hline
\multicolumn{4}{l}{\emph{Retained for dynamical analysis}}\\
U-Net H4 & 11.76 & 0.123 & 0.857\\
U-Net H4 P4 & 11.76 & 0.049 & 0.656\\
U-Net H4 K8 & 11.76 & 0.093 & 0.730\\
U-Net H1 & 11.76 & 0.037 & 0.670\\
DySLIM H1 K10 & 10.09 & 0.164 & 0.863\\
AFNO H4 & 10.29 & 0.334 & 1.106\\
AFNO H4 K2 & 10.29 & 0.076 & 0.770\\
FNO H10 K10 & 10.49 & 0.030 & 0.316\\
DPOT H10 & 7.51 & 0.141 & 0.912\\
DPOT H10 P4 & 7.51 & 0.155 & 0.883\\
PDE-Refiner H1 & 9.47 & 0.072 & 0.627\\
ACDM H2 & 11.75 & 0.289 & 0.973\\
\hline
\multicolumn{4}{l}{\emph{Not retained}}\\
PFNO H10 P10 & 10.51 & 0.619 & 6.374\\
DYffusion H1 P16 & 11.21 & 0.962 & 1.755\\
FNO H1 & 10.49 & 0.756 & 1.197\\
DPOT H10 (scratch) & 7.51 & 0.870 & 1.385\\
\hline
\multicolumn{4}{l}{%
  \textcolor{red}{\textit{Coarse DNS comparison}}}\\
\textcolor{red}{$64^2$}
& & \textcolor{red}{0.0048}
  & \textcolor{red}{0.0101}\\
\textcolor{red}{$96^2$}
& & \textcolor{red}{0.0006}
  & \textcolor{red}{0.0075}\\
\hline
\end{tabular}
\end{table}

\paragraph{Excluded configurations.}
\label{app:excluded}
PFNO H10 P10 diverged under sampled autoregressive recursion. DYffusion H1 P16, FNO H1, and DPOT H10 trained from scratch had weaker rollout performance than the retained settings. FNO H10 K10 remains in the geometry comparison but is excluded from physical-event comparisons because of its numerical instability.

\section{Derivation of the Error Decomposition}
\label{app:restart}

\subsection{Trajectory Error}
\label{app:trajectory_derivation}
{\color{red}
Following Section~\ref{sec:errors}, let $x_{k+1}=F(x_k)$, $\hat{x}_{k+1}=\hat{F}(\hat{x}_k)$, and $e_k=\hat{x}_k-x_k$. Adding and subtracting the DNS advancement $F(\hat{x}_k)$ gives
\begin{align}
e_{k+1}
&=\hat x_{k+1}-x_{k+1}\\
&=\underbrace{\bigl[\hat x_{k+1}-F(\hat x_k)\bigr]}_{r_{k+1}}
 +\underbrace{\bigl[F(\hat x_k)-x_{k+1}\bigr]}_{h_{k+1}}
=r_{k+1}+h_{k+1}.
\end{align}
Thus $r_{k+1}$ is the local prediction error of the neural update relative to DNS from the same predicted state, while $h_{k+1}=F(\hat{x}_k)-F(x_k)$ is the existing error after propagation by the DNS dynamics. Expanding the squared norm and subtracting the current squared error gives
\begin{align}
\|e_{k+1}\|_{\mathrm{sp}}^2-\|e_k\|_{\mathrm{sp}}^2
&=\|h_{k+1}+r_{k+1}\|_{\mathrm{sp}}^2-\|e_k\|_{\mathrm{sp}}^2\\
&=\underbrace{\|h_{k+1}\|_{\mathrm{sp}}^2-\|e_k\|_{\mathrm{sp}}^2}_{g_{k+1}}
+\underbrace{\|r_{k+1}\|_{\mathrm{sp}}^2}_{j_{k+1}}
+\underbrace{2\langle h_{k+1},r_{k+1}\rangle_{\mathrm{sp}}}_{c_{k+1}}.
\end{align}
Here $\langle u,v\rangle_{\mathrm{sp}}=D^{-1}\sum_i u_i v_i$ and $\|u\|_{\mathrm{sp}}^2=\langle u,u\rangle_{\mathrm{sp}}$. The identity requires neither linearization nor independence of $h_{k+1}$ and $r_{k+1}$. In particular, the nonnegative local-error term $j_{k+1}$ can be obscured by negative $g_{k+1}$ or $c_{k+1}$.

Summing over a window from $a$ to $b$ yields
\begin{equation}
\|e_b\|_{\mathrm{sp}}^2-\|e_a\|_{\mathrm{sp}}^2
=\sum_{i=a+1}^{b}g_i+\sum_{i=a+1}^{b}j_i+\sum_{i=a+1}^{b}c_i.
\end{equation}
For the rollout from the common forecast origin, $e_0=0$, so summing through step $k$ recovers the main-text identity $\|e_k\|_{\mathrm{sp}}^2=G_k+J_k+C_k$, with $G_k=\sum_{i=1}^k g_i$ and analogously for $J_k$ and $C_k$. Ensemble- or case-level summaries average these memberwise quantities only after the identity is formed.
}

For Figure~\ref{fig:budget}(c--d), a trajectory is classified as later-failing if its peak absolute vorticity exceeds ten times the maximum peak absolute vorticity of the aligned DNS ensemble for case 4232. The comparison uses all 50 threshold-crossing members and a fixed control group of 50 non-crossing members.

\subsection{Residual Effects on Ensemble Evolution}
\label{app:ensemble_derivation}
{\color{red}
Let $\mathbf r_{k+1}=\hat{\mathbf x}_{k+1}-F(\hat{\mathbf x}_k)$ be the memberwise local prediction error, and recall that the centering operator $\mathcal Z$ is linear. Define
\begin{equation}
Y_{k+1}=\mathcal ZF(\hat{\mathbf x}_k),\qquad
X_{k+1}=\mathcal Z\mathbf x_{k+1},\qquad
R_{k+1}=\mathcal Z\mathbf r_{k+1}.
\end{equation}
Then $\mathcal Z\hat{\mathbf x}_{k+1}=Y_{k+1}+R_{k+1}$. Expanding the two differences used to define $D$ and $T$ in Section~\ref{sec:errors} gives
\begin{align}
D
&=\sum_{k\in\mathcal W}\left(
\|R_{k+1}\|_M^2+2\langle Y_{k+1},R_{k+1}\rangle_M\right),\\
T
&=\sum_{k\in\mathcal W}\left(
\|R_{k+1}\|_M^2+2\langle Y_{k+1}-X_{k+1},R_{k+1}\rangle_M\right),
\end{align}
where $\langle A,B\rangle_M=(MD)^{-1}\sum_{m,i}A_{m,i}B_{m,i}$. Subtracting the two expansions yields
\begin{equation}
T-D=-2\sum_{k\in\mathcal W}\langle X_{k+1},R_{k+1}\rangle_M.
\label{eq:reference_shift}
\end{equation}
The same local prediction error can therefore make $D<0$ while making $T>0$: the two metrics use the same counterfactual DNS advancement $F(\hat{\mathbf x}_k)$ but different reference vectors in the cross term. Both are sums of per-update differences over $\mathcal W$, not changes between the window endpoints, which would also include propagation between successive predicted states.
}

\section{Event and Vortex Diagnostics}
\label{app:events}

\subsection{Event Definition and Aggregate Probability Errors}
\label{app:event_definition}
An event occurs when the largest same-sign connected component of $|\omega|\geq10$, with connectivity across periodic boundaries, occupies at least 0.5\% of the domain. Positive and negative threshold masks are labeled separately using four-neighbor connectivity, with labels merged across opposite domain edges. Diagonal contact alone does not join components. Component area is its grid-point count divided by $128^2$, so the event threshold corresponds to at least 82 grid points. The indicator $I(x)$ equals one when this criterion is satisfied and zero otherwise. For each case and stored time, event probability is the fraction of ensemble members satisfying this criterion. In Table~\ref{tab:event_frequency}, Excess and Deficit average the positive and negative parts of the solver-minus-reference probability difference separately; Bias is their signed difference. Cases receive equal weight, and the 35 stored times spanning forecast steps 1--240 receive trapezoidal weights. All three quantities are normalized by the mean reference event frequency using the same weights.

\subsection{Persistent-Event Groups}
\label{app:event_groups}
The groups in case 1616 are fixed by persistent event or non-event outcomes under the reference dynamics over forecast steps 90--100. Each solver is evaluated on these same groups. For a fixed group of $N$ trajectories and $L$ stored times, the frequency reported in Table~\ref{tab:branch_frequency} is $(NL)^{-1}\sum_{m,j}I(x_m(t_j))$. This averages event occupancy equally over members and stored times, rather than counting trajectories with at least one event. It is a within-group statistic over the selected window, distinct from the time-weighted aggregate probability errors above.

\subsection{Vortex Geometry}
\label{app:vortex_geometry}
For the local comparison in Figure~\ref{fig:local_dynamics}, we characterize vortex strength, elongation, and orientation using peak absolute vorticity $V$, core axis ratio $\chi_{70}$, and accumulated major-axis rotation. The latter describes the geometry of an Eulerian vorticity region, not fluid-particle rotation.

\paragraph{Core identification and periodic coordinates.}
For each field, $V=\max_{\boldsymbol{x}}|\omega(\boldsymbol{x})|$, and $\sigma$ is the vorticity sign at that maximum. We select the largest periodic four-neighbor connected component satisfying $\sigma\omega\geq0.7V$. This is an area-based selection among components of the peak's sign; the selected component need not contain the absolute maximum itself. Before computing geometric moments, we obtain the circular mean of the core coordinates separately in each periodic direction and map each coordinate to its minimum-image offset from that mean. This avoids treating points adjacent across a periodic boundary as widely separated. The construction describes a localized core for which this periodic coordinate chart is well defined.

\paragraph{Axis ratio and orientation.}
Let $\boldsymbol{\xi}_i$ be the resulting coordinates of the $N$ core grid points and $\bar{\boldsymbol{\xi}}$ their arithmetic mean. We form the area-weighted second-moment matrix
\begin{equation}
S=\frac1N\sum_{i=1}^{N}(\boldsymbol{\xi}_i-\bar{\boldsymbol{\xi}})
(\boldsymbol{\xi}_i-\bar{\boldsymbol{\xi}})^\top,
\qquad \chi_{70}=\sqrt{\frac{\eta_1}{\eta_2}},
\end{equation}
where $\eta_1\geq\eta_2>0$ are the eigenvalues of $S$. Equal grid spacing makes grid-coordinate and physical-coordinate axis ratios identical. The eigenvector associated with $\eta_1$ defines the major-axis angle $\theta$ modulo $\pi$. All core points have equal weight; vorticity magnitudes do not weight the moments. Geometry is undefined for a zero field, fewer than three core points, or a nonpositive minor eigenvalue; non-finite fields are also excluded from this calculation.

\paragraph{Accumulated rotation.}
For successive valid orientations, we remove the sign ambiguity of the major eigenvector using
\begin{equation}
\Delta\theta_j=\tfrac12\operatorname{atan2}\!\left(
\sin[2(\theta_j-\theta_{j-1})],
\cos[2(\theta_j-\theta_{j-1})]\right).
\end{equation}
We accumulate these increments and report angles in degrees. Orientations with $\chi_{70}<1.05$ are omitted because near-circular cores have poorly determined major axes. An invalid orientation interrupts accumulation: the next valid segment starts at zero, and increments are not inferred across the gap. No such reset occurs over the interval shown in Figure~\ref{fig:local_dynamics}. Interpreting the accumulated angle as continuous rotation requires a consistently identified core and inter-frame angular changes smaller than $\pi/2$. Figure~\ref{fig:local_dynamics} reports rotation relative to the DNS median.

\section{Ensemble Geometry Diagnostics}
\label{app:geometry}
For each case $c$ and solver $q$, the centered fields are $z^q_{c,m}(t)=x^q_{c,m}(t)-M^{-1}\sum_{m'}x^q_{c,m'}(t)$. Centering separates variation across members from ensemble-mean bias.

{\color{red}
Let $Z_c^q(t)\in\mathbb R^{D\times M}$ contain the centered members as columns. Consistent with the population normalization in the main text, we use
\begin{equation}
C_c^q(t)=\frac{Z_c^q(t)Z_c^q(t)^\top}{DM}.
\end{equation}
Its Euclidean-orthonormal eigenvectors are the POD modes, and its eigenvalues satisfy $\sum_i\lambda_{c,i}^q(t)=M^{-1}\sum_m\|z^q_{c,m}(t)\|_{\mathrm{sp}}^2=[S_c^q(t)]^2$. Thus spread is the square root of total mode variance, not the RMS of the eigenvalues. The spread and 90\%-effective rank are
\begin{equation}
\begin{gathered}
S_c^q(t)=\left[\frac1M\sum_m\|z^q_{c,m}(t)\|_{\mathrm{sp}}^2\right]^{1/2},\qquad
r_{90,c}^q(t)=\min\left\{r:\frac{\sum_{i=1}^r\lambda_{c,i}^q(t)}{\sum_i\lambda_{c,i}^q(t)}\geq0.9\right\}.
\end{gathered}
\label{eq:spread_dimension}
\end{equation}
Let $U_c^q(t)$ contain the leading $r_{90,c}^q(t)$ POD modes. The main-text temporal reorientation is
\begin{equation}
R_c^q(t)=1-\frac{1}{r_{90,c}^q(t)}
\left\|U_c^q(t)^\top U_c^q(t+10)\right\|_F^2.
\label{eq:temporal_reorientation}
\end{equation}
This definition uses each time's effective-rank subspace, rather than a fixed rank of three. It is zero when the subspace at $t$ is fully contained in the comparison subspace at $t+10$, and increases as their normalized overlap decreases. Figure~\ref{fig:geometry} normalizes spread by the DNS case-median spread at the first stored time and reports case medians of $S$, $r_{90}$, and $R$.
}

{\color{red}
For solver--DNS comparison, let $r_{\min,c}(t)=\min(r_{90,c}^{\mathrm{solver}}(t),r_{90,c}^{\mathrm{DNS}}(t))$ and truncate both bases to this common dimension. Their overlap is
\begin{equation}
O_{r_{\min},c}^{\mathrm{solver,DNS}}(t)
=\frac{1}{r_{\min,c}(t)}
\left\|U_{r_{\min},c}^{\mathrm{solver}}(t)^\top
U_{r_{\min},c}^{\mathrm{DNS}}(t)\right\|_F^2.
\label{eq:subspace_overlap}
\end{equation}
}
Comparing subspaces rather than individual POD modes avoids interpreting a change of basis within a shared subspace as physical disagreement. Figure~\ref{fig:overlap} reports case-median overlaps and comparison ranks at each step; its summary over steps 50--90 first takes the temporal median within each case and then reports the median and 10th--90th percentiles across cases.
\clearpage
{\color{red}
\section{Sensitivity to Model-Noise Coupling}
\label{app:noise_coupling}

\subsection{Protocols and Evaluation Scope}

We compare the main shared-path protocol (S), which conditions each case on one model-noise realization, with member-independent paths (I), which also sample model randomness across members. Both draw fresh noise at successive updates; one trajectory is generated per initial state. PDE-Refiner H1 uses B48/M256 and ACDM H2 uses a matched B12/M32 subset, with member 0 excluded. Error probes cover steps 0--80 and geometry/events use 35 saved times over steps 1--240. Comparisons are within model because the cohorts differ.

\subsection{Rollout Error and Error Formation}

Using Section~\ref{sec:errors} and Appendix~\ref{app:restart}, Figure~\ref{fig:noise_error_formation} reports equal-case means after member averaging. At step 80, PDE-Refiner error is 2.910 under S and 2.794 under I, with $G$ accounting for 97.16\% and 97.33\%. For ACDM, I is worse at steps 20 and 40 but slightly better at step 80 (7.619 versus 7.790), where $C$ changes from $+6.95\%$ to $-1.84\%$. Noise coupling therefore changes both error and its formation without a uniform benefit. The audited restart remainder satisfies $\max |Q_k|/E_k^2<7.0\times10^{-5}$ and is not plotted.

\begin{figure}[!htbp]
\centering
\color{red}
\includegraphics[width=\linewidth]{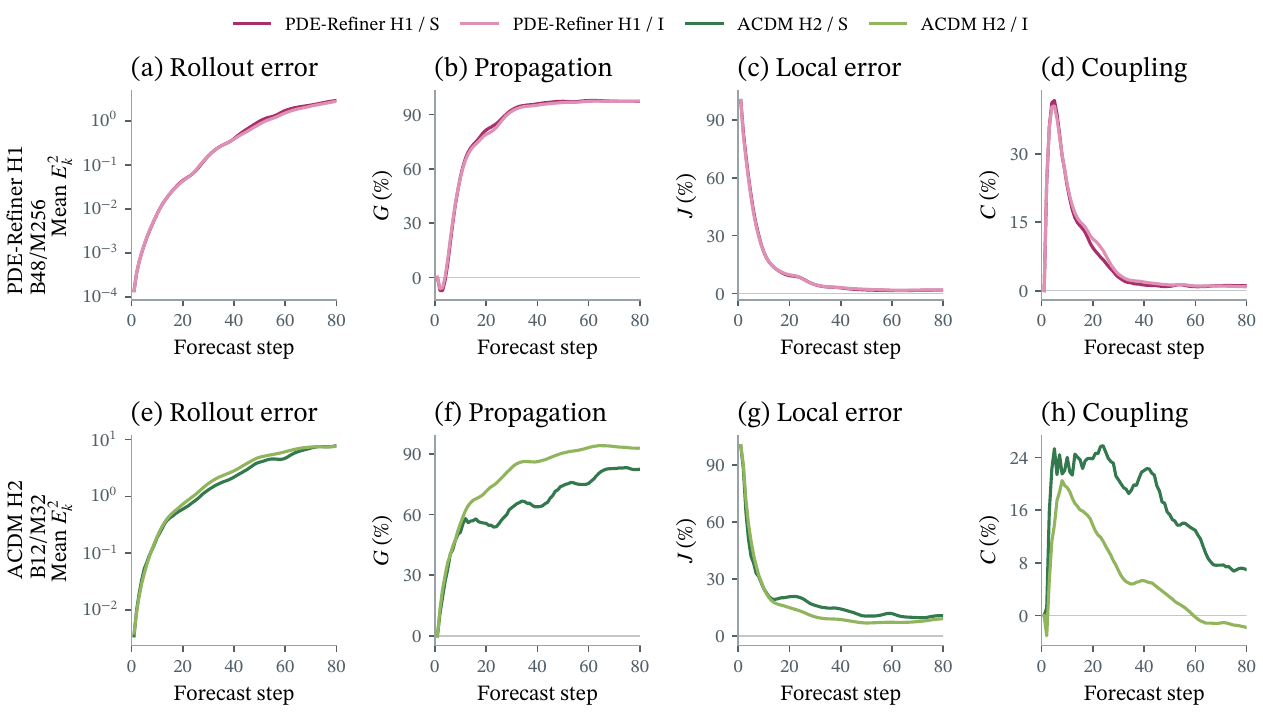}
\caption{Rollout error and cumulative $G/J/C$ under shared (S) and member-independent (I) noise. Rows show PDE-Refiner H1 (B48/M256) and ACDM H2 (B12/M32). Contributions are percentages of mean squared error after member and equal-case averaging; $Q$ is not plotted.}
\label{fig:noise_error_formation}
\end{figure}

\subsection{Relative Ensemble-Size Change and Deviation}

For ACDM, we evaluate the cumulative $D$ and $T$ from Section~\ref{sec:errors}; conversion to the stored-DNS reference changes tested windows by at most $2.93\times10^{-7}$ without changing signs. Over steps 0--32, S has $D<0$ in all 12 cases and $D<0,T>0$ in nine, whereas I has only three cases with $D<0$ and none with both signs. At step 32, the median solver/DNS spread ratio is 0.81 under S versus 10.58 under I. Over steps 32--64, all cases have $T<0$, while $D<0$ in 12 S and 11 I cases. These are update-relative effects, not endpoint changes or probability-law recovery.

\begin{figure}[!htbp]
\centering
\color{red}
\includegraphics[width=\linewidth]{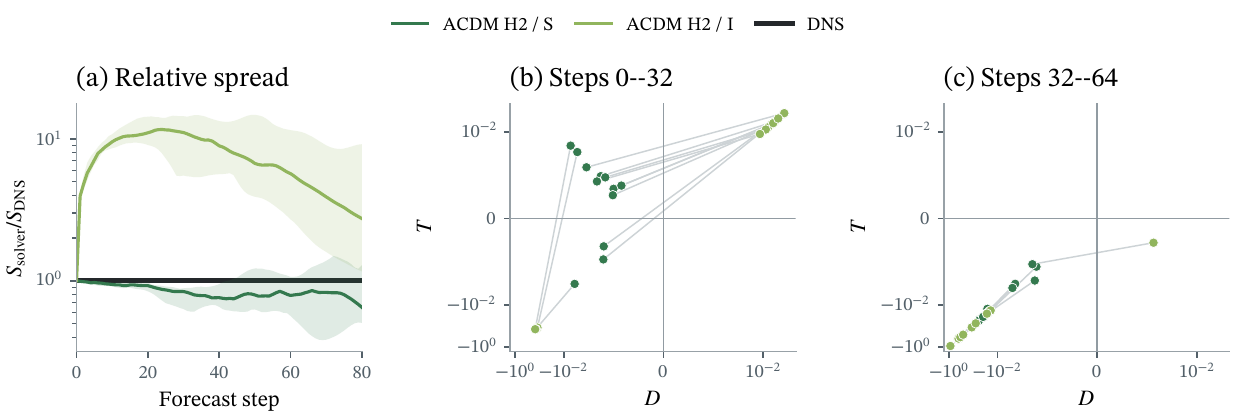}
\caption{ACDM H2 on B12/M32. (a) Solver/DNS RMS-spread ratio (median and 10th--90th case percentiles). (b,c) Cumulative $D,T$ over $[0,32)$ and $[32,64)$; points are cases and gray segments pair S with I. Scatter axes are symmetric-log with linear scale $10^{-5}$.}
\label{fig:noise_acdm_effects}
\end{figure}

\begin{table}[!htbp]
\centering
\color{red}
\small
\begin{tabular}{lrrrr}
\hline
Protocol & Window & $D<0$ & $T>0$ & Both \\
\hline
S & 0--32 & 12 & 9 & 9 \\
S & 32--64 & 12 & 0 & 0 \\
I & 0--32 & 3 & 9 & 0 \\
I & 32--64 & 11 & 0 & 0 \\
\hline
\end{tabular}

\caption{ACDM sign counts out of 12 cases; ``Both'' means $D<0$ and $T>0$.}
\label{tab:noise_dt_counts}
\end{table}

\subsection{Spread, Dimension, and DNS Alignment}

Figure~\ref{fig:noise_geometry} uses the definitions in Section~\ref{sec:geometry} and Appendix~\ref{app:geometry}. Median overlap over steps 50--90 increases from 0.301 to 0.397 for PDE-Refiner and from 0.026 to 0.067 for ACDM under I, but ACDM remains poorly aligned. These realized-trajectory comparisons are descriptive and do not isolate noise coupling causally.

\begin{figure}[!htbp]
\centering
\color{red}
\includegraphics[width=\linewidth]{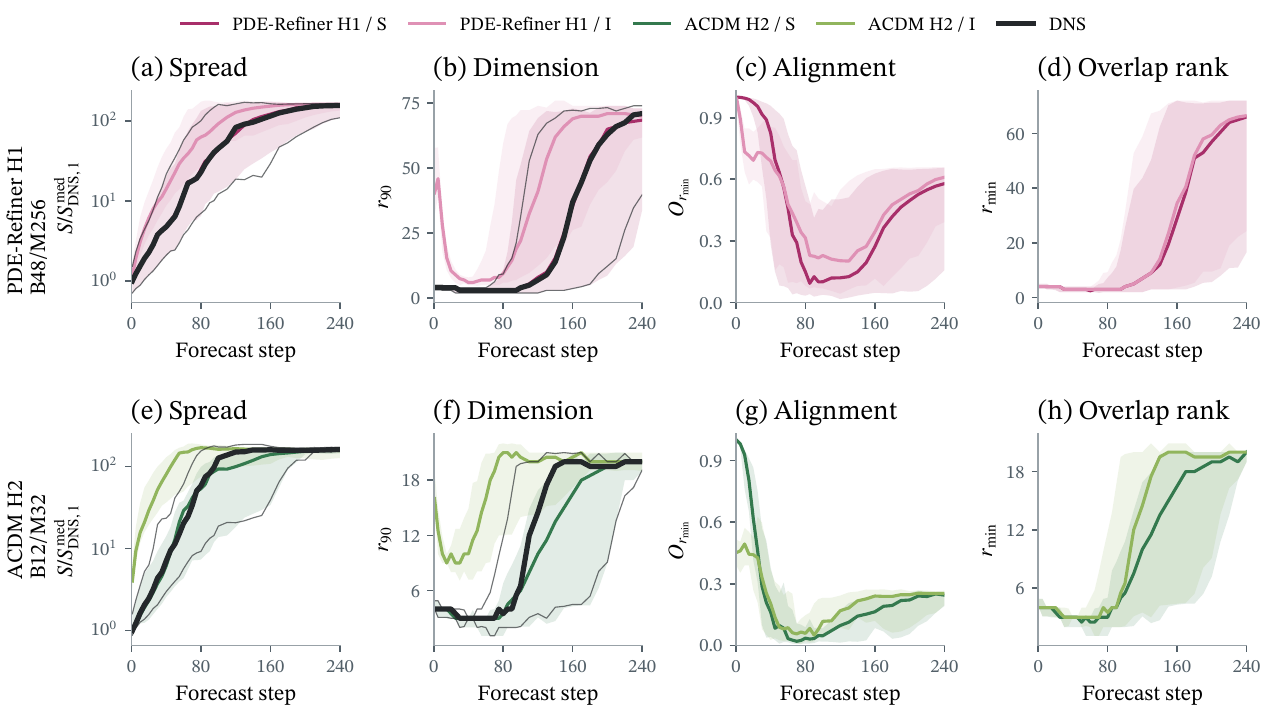}
\caption{Ensemble geometry under S and I. Rows show PDE-Refiner H1 (B48/M256) and ACDM H2 (B12/M32); columns show normalized spread, $r_{90}$, solver--DNS overlap, and $r_{\min}$. Curves are case medians with 10th--90th percentiles; black curves denote DNS.}
\label{fig:noise_geometry}
\end{figure}

\subsection{Casewise Extreme-Event Probabilities}

Using Appendix~\ref{app:event_definition}, PDE-Refiner has frequency bias $+0.44\%$ under S and $+4.74\%$ under I, but excess/deficit remain 36.74/36.30\% and 41.38/36.64\%. Aggregate agreement therefore hides case/time errors under both protocols. On the ACDM subset, bias worsens from $-47.02\%$ to $-64.48\%$; its matched DNS frequency is 10.65\%, so these values cannot replace the full-cohort ACDM scores.

\begin{figure}[!htbp]
\centering
\color{red}
\includegraphics[width=\linewidth]{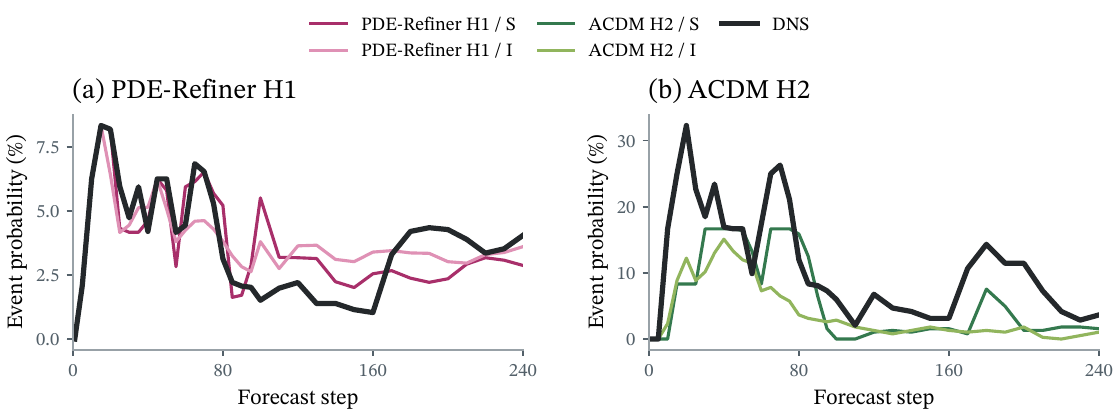}
\caption{Case-averaged event probability under S and I with matched DNS cohorts. Panels use different cohorts and scales. Table~\ref{tab:noise_event_frequency} adds time weighting and separates excess from deficit.}
\label{fig:noise_event_probability}
\end{figure}

\begin{table}[!htbp]
\centering
\color{red}
\small
\begin{tabular}{lrrrrr}
\hline
Model / protocol & DNS (\%) & Solver (\%) & Bias (\%) & Excess (\%) & Deficit (\%) \\
\hline
PDE-Refiner H1 / S & 3.62 & 3.64 & 0.44 & 36.74 & 36.30 \\
PDE-Refiner H1 / I & 3.62 & 3.79 & 4.74 & 41.38 & 36.64 \\
ACDM H2 / S & 10.65 & 5.64 & -47.02 & 7.82 & 54.85 \\
ACDM H2 / I & 10.65 & 3.78 & -64.48 & 10.13 & 74.61 \\
\hline
\end{tabular}

\caption{Time-weighted event frequencies and errors. DNS/Solver are probability percentages; Bias/Excess/Deficit are percentages of matched DNS frequency.}
\label{tab:noise_event_frequency}
\end{table}

\subsection{Interpretation Limits}

I includes model-noise variability and has no repeated-noise uncertainty estimate. ACDM S and I also differ in hardware, PyTorch version, and padded batch size. The comparison therefore establishes protocol sensitivity, not a causal effect of noise coupling. PDE-Refiner $D/T$ and independent-noise case 1616 are unavailable.
}

\clearpage
\providecommand{\allocationroot}{data_allocation_appendix}
{\color{red}
\section{Training-data Allocation at a Fixed Trajectory Count}
\label{app:data_allocation}

\subsection{Experimental setting and evaluation}
We redistribute 416 training trajectories among 416$\times$1 unperturbed baselines, 104$\times$4 perturbed members, or 26$\times$16 perturbed members (250 frames each; member 0 excluded in the controls). This changes case coverage and trajectory type, not total trajectory count.

We train U-Net H4, AFNO H4, FNO H10 K10, and PDE-Refiner H1 with their original initialization, schedule, effective batch size, and update count (42,640, 19,950, 30,000, and 20,200). The same disjoint 48-case validation and test sets are used. Each allocation has one training realization; controls use PyTorch 2.9 rather than 2.5.1, so results are descriptive rather than seed-averaged causal effects.

Forecast step $k=0$ is simulation step 10. Rollout, geometry, and events use B48/M256; error formation uses the matched M8 cohort through $k=80$. PDE-Refiner uses stored shared-noise rollouts. Metrics follow Sections~\ref{sec:errors}--\ref{sec:geometry} and Appendices~\ref{app:restart} and~\ref{app:geometry}. The plotted spread ratio $S^{\mathrm{solver}}/S^{\mathrm{DNS}}$ is invariant to the archived common $M-1$ factor; unity means equal spread, not aligned subspaces.

\subsection{Rollout error and ensemble geometry}
Allocation effects are solver-dependent (Figure~\ref{fig:allocation_diagnostics}; Table~\ref{tab:allocation_geometry}). Both controls increase median error for U-Net and AFNO, while PDE-Refiner improves error and overlap. FNO 104$\times$4 improves overlap despite higher error. PDE-Refiner 26$\times$16 improves case-level error in 26/48 cases and overlap in 30/48.

\begin{figure}[!htbp]
\centering
\color{red}
\includegraphics[width=\linewidth]{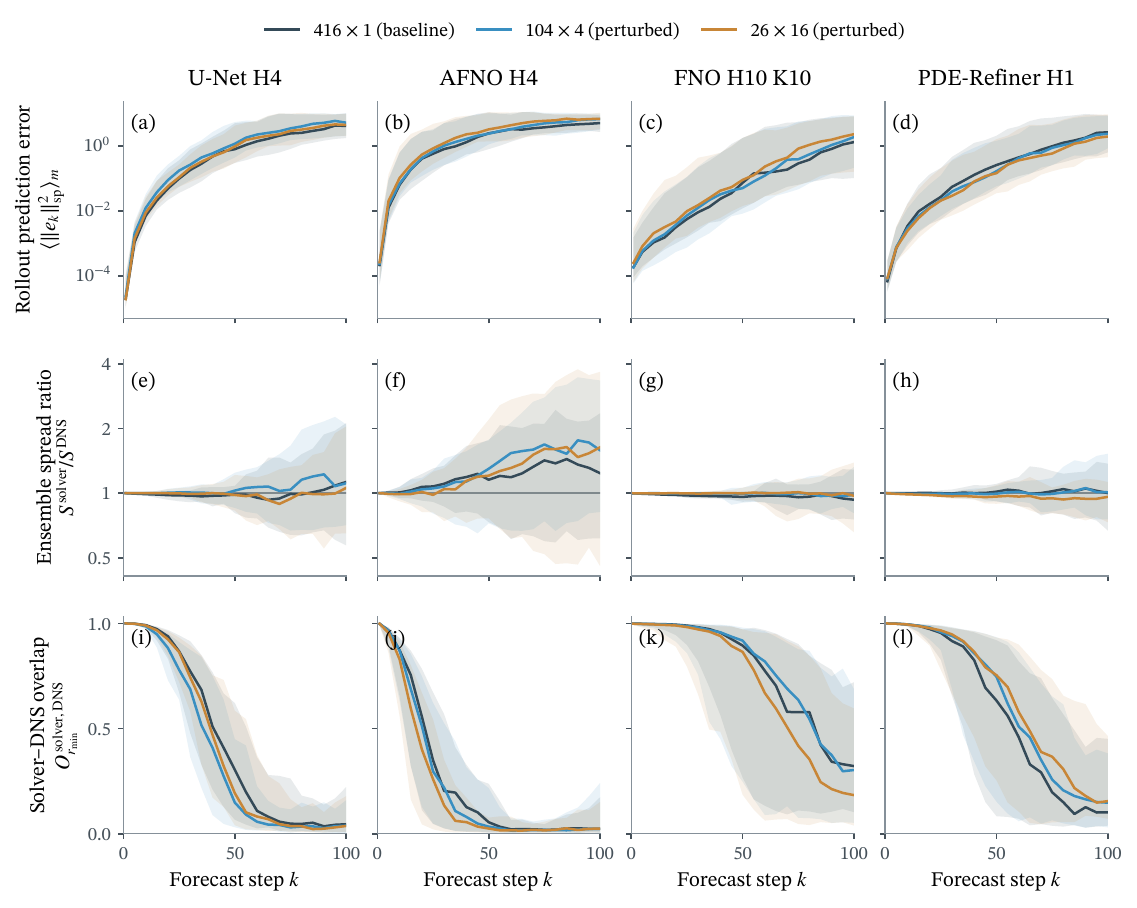}
\caption{Allocation effects on rollout error, solver/DNS spread ratio, and adaptive-rank overlap for four solvers. Curves show B48/M256 case medians with 10th--90th percentiles. The legend gives training cases $\times$ trajectories per case; 416$\times$1 uses baselines and the controls use perturbed members. Panels show steps 1--100; full 1--240 data, including the FNO late tail, are retained separately.}
\label{fig:allocation_diagnostics}
\end{figure}

\begin{table}[!htbp]
\centering\color{red}\small\setlength{\tabcolsep}{4pt}
\caption{B48/M256 diagnostics: temporal medians over steps 50--90 within case, then medians across cases. Error is member-averaged spatial MSE; overlap is evaluated at $r_{\min}$.}
\label{tab:allocation_geometry}
\begin{tabular}{llrrrrr}
\hline
Solver & Allocation & Rollout error & $S^{\rm solver}/S^{\rm DNS}$ & $r_{90}$ & $r_{\min}$ & $O_{r_{\min}}$ \\
\hline
U-Net H4 & 416$\times$1 & 1.895 & 0.952 & 3 & 3 & 0.073 \\
U-Net H4 & 104$\times$4 & 2.857 & 1.057 & 3 & 3 & 0.039 \\
U-Net H4 & 26$\times$16 & 2.298 & 0.970 & 3 & 3 & 0.049 \\
\hline
AFNO H4 & 416$\times$1 & 3.544 & 1.296 & 3 & 3 & 0.025 \\
AFNO H4 & 104$\times$4 & 4.348 & 1.577 & 3 & 3 & 0.018 \\
AFNO H4 & 26$\times$16 & 5.524 & 1.525 & 3 & 2 & 0.018 \\
\hline
FNO H10 K10 & 416$\times$1 & 0.201 & 0.966 & 3 & 3 & 0.616 \\
FNO H10 K10 & 104$\times$4 & 0.348 & 0.994 & 3 & 3 & 0.690 \\
FNO H10 K10 & 26$\times$16 & 0.437 & 1.007 & 3 & 3 & 0.531 \\
\hline
PDE-Refiner H1 & 416$\times$1 & 0.640 & 1.019 & 3 & 3 & 0.301 \\
PDE-Refiner H1 & 104$\times$4 & 0.620 & 1.002 & 3 & 3 & 0.359 \\
PDE-Refiner H1 & 26$\times$16 & 0.505 & 0.951 & 3 & 3 & 0.416 \\
\hline
\end{tabular}

\end{table}

\subsection{DNS-probed error formation}
We use the memberwise $G_k/J_k/C_k$ decomposition in Appendix~\ref{app:restart}, then average over M8 and cases. A restart correction $Q_k$ retains numerical closure: its maximum cumulative magnitude is 0.0276\% of error, and the maximum algebraic/Parseval residual is $2.25\times10^{-15}$.

At $k=20$, U-Net 26$\times$16 matches the original error despite 63\% larger $J_k$ and 64\% larger $C_k$, offset by lower $G_k$; both controls increase $J_k$ in all 48 cases at $k=20,40$. At $k=80$, PDE-Refiner 26$\times$16 has 23\% lower error despite 15\% larger $J_k$ ($\Delta G=-0.65505$, $\Delta J=+0.00710$, $\Delta C=-0.04194$), with lower error and larger $J_k$ in 24/48 cases. AFNO 104$\times$4 instead has smaller $J_k$ but larger error because $G_k$ increases and $C_k$ is less negative.

These descriptive decompositions are not interventions. M8 error formation and M256 geometry also use different cohorts and aggregation rules.

\begin{table}[!htbp]
\centering\color{red}\small\setlength{\tabcolsep}{4pt}
\caption{Error formation at $k=20$, averaged over M8 and then 48 cases. $G_k/J_k/C_k$ are cumulative and use squared-error units; audited $Q_k$ is omitted.}
\label{tab:allocation_formation20}
\begin{tabular}{llrrrr}
\hline
Solver & Allocation & $\langle\|e_k\|_{\rm sp}^2\rangle_{c,m}$ & $\langle G_k\rangle_{c,m}$ & $\langle J_k\rangle_{c,m}$ & $\langle C_k\rangle_{c,m}$ \\
\hline
U-Net H4 & 416$\times$1 & 0.08105 & 0.06756 & 0.00238 & 0.01111 \\
U-Net H4 & 104$\times$4 & 0.13057 & 0.10170 & 0.00436 & 0.02452 \\
U-Net H4 & 26$\times$16 & 0.08109 & 0.05895 & 0.00388 & 0.01827 \\
\hline
AFNO H4 & 416$\times$1 & 0.90735 & 0.58398 & 0.16207 & 0.16130 \\
AFNO H4 & 104$\times$4 & 0.89264 & 0.50593 & 0.17088 & 0.21583 \\
AFNO H4 & 26$\times$16 & 1.17014 & 0.64377 & 0.20706 & 0.31931 \\
\hline
FNO H10 K10 & 416$\times$1 & 0.01838 & 0.00239 & 0.01425 & 0.00175 \\
FNO H10 K10 & 104$\times$4 & 0.02761 & 0.00598 & 0.01780 & 0.00383 \\
FNO H10 K10 & 26$\times$16 & 0.04815 & 0.01420 & 0.02395 & 0.01001 \\
\hline
PDE-Refiner H1 & 416$\times$1 & 0.04415 & 0.03592 & 0.00408 & 0.00414 \\
PDE-Refiner H1 & 104$\times$4 & 0.04164 & 0.03198 & 0.00487 & 0.00480 \\
PDE-Refiner H1 & 26$\times$16 & 0.03814 & 0.02485 & 0.00687 & 0.00642 \\
\hline
\end{tabular}

\end{table}
\begin{table}[!htbp]
\centering\color{red}\small\setlength{\tabcolsep}{4pt}
\caption{Error formation at $k=80$, using the same definitions and averaging as Table~\ref{tab:allocation_formation20}.}
\label{tab:allocation_formation80}
\begin{tabular}{llrrrr}
\hline
Solver & Allocation & $\langle\|e_k\|_{\rm sp}^2\rangle_{c,m}$ & $\langle G_k\rangle_{c,m}$ & $\langle J_k\rangle_{c,m}$ & $\langle C_k\rangle_{c,m}$ \\
\hline
U-Net H4 & 416$\times$1 & 4.10246 & 4.12527 & 0.01448 & -0.03732 \\
U-Net H4 & 104$\times$4 & 5.00281 & 4.81484 & 0.02444 & 0.16351 \\
U-Net H4 & 26$\times$16 & 4.45554 & 4.34376 & 0.01813 & 0.09362 \\
\hline
AFNO H4 & 416$\times$1 & 5.41223 & 5.32513 & 0.82226 & -0.73520 \\
AFNO H4 & 104$\times$4 & 6.15518 & 5.87765 & 0.78470 & -0.50720 \\
AFNO H4 & 26$\times$16 & 6.25621 & 6.18243 & 0.85170 & -0.77796 \\
\hline
FNO H10 K10 & 416$\times$1 & 2.05752 & 2.03131 & 0.10925 & -0.08305 \\
FNO H10 K10 & 104$\times$4 & 2.02320 & 1.97939 & 0.09914 & -0.05534 \\
FNO H10 K10 & 26$\times$16 & 2.16013 & 2.09865 & 0.09752 & -0.03604 \\
\hline
PDE-Refiner H1 & 416$\times$1 & 2.95379 & 2.87611 & 0.04876 & 0.02891 \\
PDE-Refiner H1 & 104$\times$4 & 2.40017 & 2.35368 & 0.06395 & -0.01746 \\
PDE-Refiner H1 & 26$\times$16 & 2.26390 & 2.22106 & 0.05587 & -0.01303 \\
\hline
\end{tabular}

\end{table}

\subsection{Conditional event probabilities}
We use Appendix~\ref{app:event_definition} with B48/M256 and steps 1--240. Bias/Excess/Deficit are percentages of the matched DNS frequency (3.6201\%), whereas the table's Frequency column is an ordinary probability percentage.

U-Net controls retain strong underprediction; FNO controls increase Excess and total probability error. PDE-Refiner 26$\times$16 slightly reduces total error while moving aggregate frequency farther from DNS, again separating frequency agreement from conditional accuracy.

\begin{table}[!htbp]
\centering\color{red}\small\setlength{\tabcolsep}{4pt}
\caption{B48/M256 event statistics over steps 1--240. Frequency is an ordinary probability percentage; Bias/Excess/Deficit are normalized by DNS frequency. PDE-Refiner uses shared noise.}
\label{tab:allocation_events}
\begin{tabular}{llrrrr}
\hline
Solver & Allocation & Frequency (\%) & Bias (\%) & Excess (\%) & Deficit (\%) \\
\hline
U-Net H4 & 416$\times$1 & 1.435 & -60.36 & 11.08 & 71.44 \\
U-Net H4 & 104$\times$4 & 1.512 & -58.22 & 26.75 & 84.97 \\
U-Net H4 & 26$\times$16 & 1.366 & -62.28 & 20.73 & 83.01 \\
\hline
AFNO H4 & 416$\times$1 & 3.369 & -6.93 & 53.42 & 60.35 \\
AFNO H4 & 104$\times$4 & 3.138 & -13.32 & 47.23 & 60.55 \\
AFNO H4 & 26$\times$16 & 4.074 & +12.53 & 80.72 & 68.19 \\
\hline
FNO H10 K10 & 416$\times$1 & 4.154 & +14.75 & 44.26 & 29.51 \\
FNO H10 K10 & 104$\times$4 & 6.071 & +67.69 & 85.06 & 17.37 \\
FNO H10 K10 & 26$\times$16 & 5.219 & +44.17 & 64.96 & 20.79 \\
\hline
PDE-Refiner H1 & 416$\times$1 & 3.636 & +0.44 & 36.74 & 36.30 \\
PDE-Refiner H1 & 104$\times$4 & 5.053 & +39.58 & 68.69 & 29.11 \\
PDE-Refiner H1 & 26$\times$16 & 4.090 & +12.98 & 41.58 & 28.60 \\
\hline
\end{tabular}

\end{table}

\paragraph{Scope.}
These controls redistribute fixed trajectory count and do not establish a universally better allocation. Lower rollout error still need not imply smaller cumulative local error, while spread agreement and aggregate event frequency remain insufficient for dynamical fidelity. We make no temporal-reorientation, centered-cloud residual-effect, or network-intervention claim.
}

\end{document}